\documentclass{article}
\usepackage{ijcai24}

\usepackage{times}
\usepackage{soul}
\usepackage{url}
\usepackage[hidelinks]{hyperref}
\usepackage[utf8]{inputenc}
\usepackage[small]{caption}
\usepackage{graphicx}
\usepackage{amsmath}
\usepackage{amsthm}
\usepackage{booktabs}
\usepackage{algorithm}
\usepackage{algorithmic}
\usepackage{multirow}
\usepackage{amsfonts}
\usepackage[switch]{lineno}
\usepackage{bm}
\usepackage{amssymb}
\usepackage{pifont}
\usepackage{bbding}
\usepackage{multirow}
\usepackage{graphicx}  
\usepackage{subcaption}  
\usepackage{pgfplots}
\usepackage{listings}
\pgfplotsset{
    every axis label/.style={font=\small},
    tick label style={font=\footnotesize}
}
\pgfplotsset{compat=1.18} 
\usepackage{xcolor}
\definecolor{maroon}{rgb}{0.5, 0.0, 0.0} 
\title{Automated Detection of Pre-training Text in Black-box LLMs\thanks{Code: \href{https://github.com/STAIR-BUPT/VeilProbe}{\texttt{github.com/STAIR-BUPT/VeilProbe}}}}

\author{
Ruihan Hu$^1$
\and
Yu-Ming Shang$^1$\and
Jiankun Peng$^{1}$\and
Wei Luo$^1$\and
Yazhe Wang$^2$\And
Xi Zhang$^1$\thanks{Corresponding Author}\\
\affiliations
$^1$Key Laboratory of Trustworthy Distributed Computing and Service (MoE),
\\Beijing University of Posts and Telecommunications, China\\
$^2$Zhongguancun Laboratory, China\\
\emails
\{gloria-1019, shangym, projankt, luowei\}@bupt.edu.cn,
wangyz@zgclab.edu.cn,
zhangx@bupt.edu.cn
}
\begin{document}

\maketitle

\begin{abstract}
Detecting whether a given text is a member of the pre-training data of Large Language Models (LLMs) is crucial for ensuring data privacy and copyright protection. Most existing methods rely on the LLM's hidden information (e.g., model parameters or token probabilities), making them ineffective in the black-box setting, where only input and output texts are accessible. Although some methods have been proposed for the black-box setting, they rely on massive manual efforts such as designing complicated questions or instructions. To address these issues, we propose VeilProbe, the first framework for automatically detecting LLMs' pre-training texts in a black-box setting without human intervention. VeilProbe utilizes a sequence-to-sequence mapping model to infer the latent mapping feature between the input text and the corresponding output suffix generated by the LLM. Then it performs the key token perturbations to obtain more distinguishable membership features. Additionally, considering real-world scenarios where the ground-truth training text samples are limited, a prototype-based membership classifier is introduced to alleviate the overfitting issue. Extensive evaluations on three widely used datasets demonstrate that our framework is effective and superior in the black-box setting.
\end{abstract}

\section{Introduction}

The extensive corpus used during the pre-training phase is a major factor behind the extraordinary performance of Large Language Models (LLMs)~\cite{NEURIPS2020_1457c0d6}.
In fact, some corpora potentially include sensitive information, such as copyrighted materials. This could give rise to legal disputes, such as The New York Times lawsuit against OpenAI~\cite{grynbaum2023nyt} for unauthorized use of its articles in training LLMs.
Pre-training data detection task in LLMs aims to determine whether a given material is included in the target LLM’s (\textit{i.e.}, the LLM to be detected) pre-training corpus~\cite{shi2024detecting}. It has become a research hotspot due to its important value in addressing issues such as detecting data contamination~\cite{oren2023provingtestsetcontamination,ye2024datacontaminationcalibrationblackbox} and protecting the rights of copyright holders~\cite{karamolegkou2023copyrightviolationslargelanguage,chang2023speakmemoryarchaeologybooks}. 

\begin{figure}[t]
  \includegraphics[width=1\columnwidth]{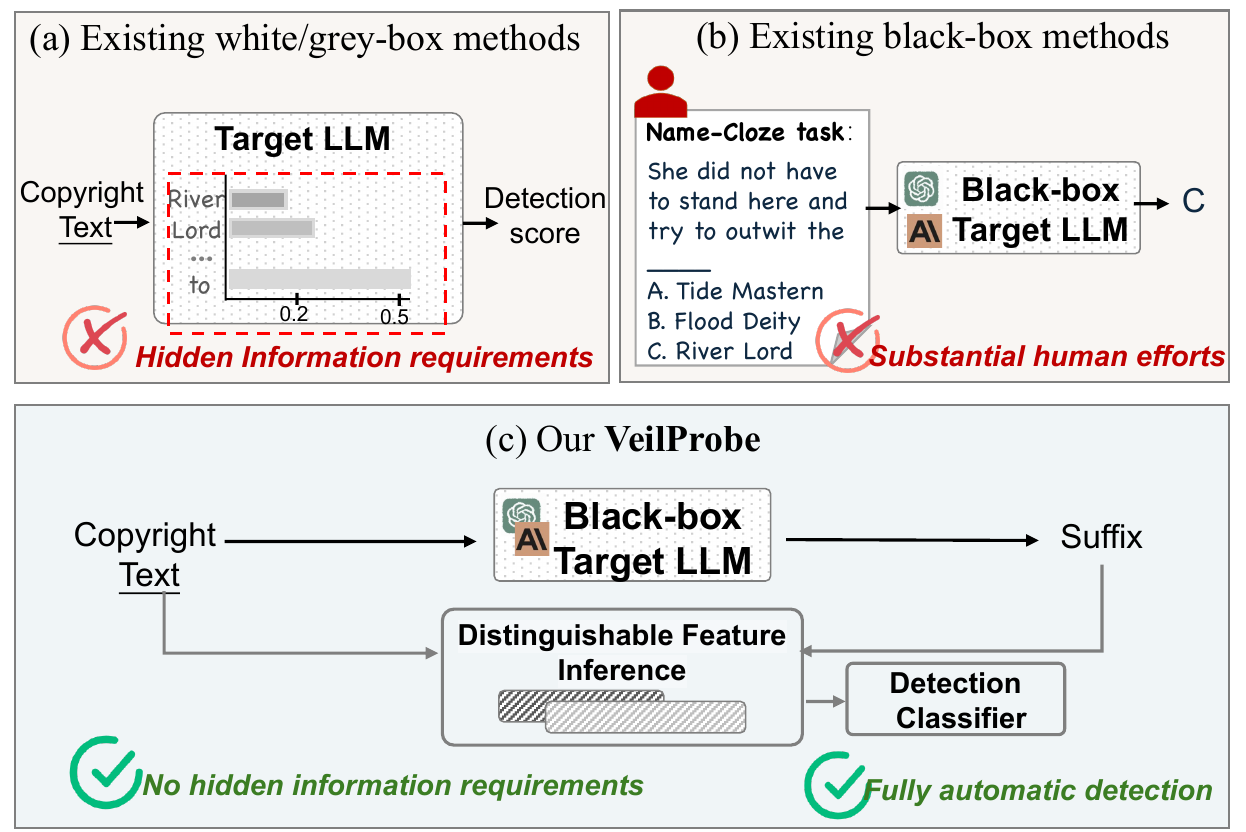}
  \caption{The comparation between existing methods and \textbf{VeilProbe (ours)}. (a) Existing white/grey-box methods should rely on hidden information of LLMs to calculate distinguishable scores for detection. (b) Exiting black-box methods requires substantial human efforts to design specific tasks for each text. (c) Our VeilProbe relies solely on text-to-suffix pairs to achieve automatic detection. }
  \label{fig: intro}
\end{figure}

Pre-training detection methods typically rely on the target LLM's distinct output behavior when the given pre-training texts are used as input. According to the detector's access levels, they can be categorized as the white-box~\cite{wang2024pandoraswhiteboxprecisetraining}, grey-box~\cite{shi2024detecting,zhang2024pretrainingdatadetectionlarge}, and black-box settings~\cite{duarte2024decopdetectingcopyrightedcontent,chang2023speakmemoryarchaeologybooks}. The methods in white-box and grey-box settings, as illustrated in Fig.~\ref{fig: intro} (a), generally follow a paradigm that calculates a distinguishable score based on the statistical characteristics of token probabilities or hidden parameters~\cite{carlini2021extracting,shi2024detecting,zhang2024pretrainingdatadetectionlarge}. For instance, Min-K\% Prob~\cite{shi2024detecting} uses the sum of the lowest token log probabilities as the score.  DC-PDD~\cite{zhang2024pretrainingdatadetectionlarge} introduces a calibrated score based on the cross-entropy between token probabilities and reference corpus frequencies. Nevertheless, advanced LLMs like ChatGPT and Claude are often in the black-box settings, where only the input and output texts are accessible. White-box and grey-box methods are infeasible in this scenario.  

Existing black-box pre-training data detection methods typically enable the target LLM to perform complicated tasks such as question-answering tasks ~\cite{karamolegkou2023copyrightviolationslargelanguage}, multiple-choice questions~\cite{duarte2024decopdetectingcopyrightedcontent}, and name-cloze tasks~\cite{chang2023speakmemoryarchaeologybooks}. The LLM's responses distinguish between pre-training texts and non-training texts. Fig.~\ref{fig: intro} (b) shows an example that the name-cloze task requires deciding where to place blanks and creating customized options. Yet, it requires handcrafted instructions for each specific input, which is expensive and time-consuming, and cannot adapt to real-world large-scale pre-training text detection.

\begin{figure*}[t]
  \centering  
  \includegraphics[width=1\textwidth]{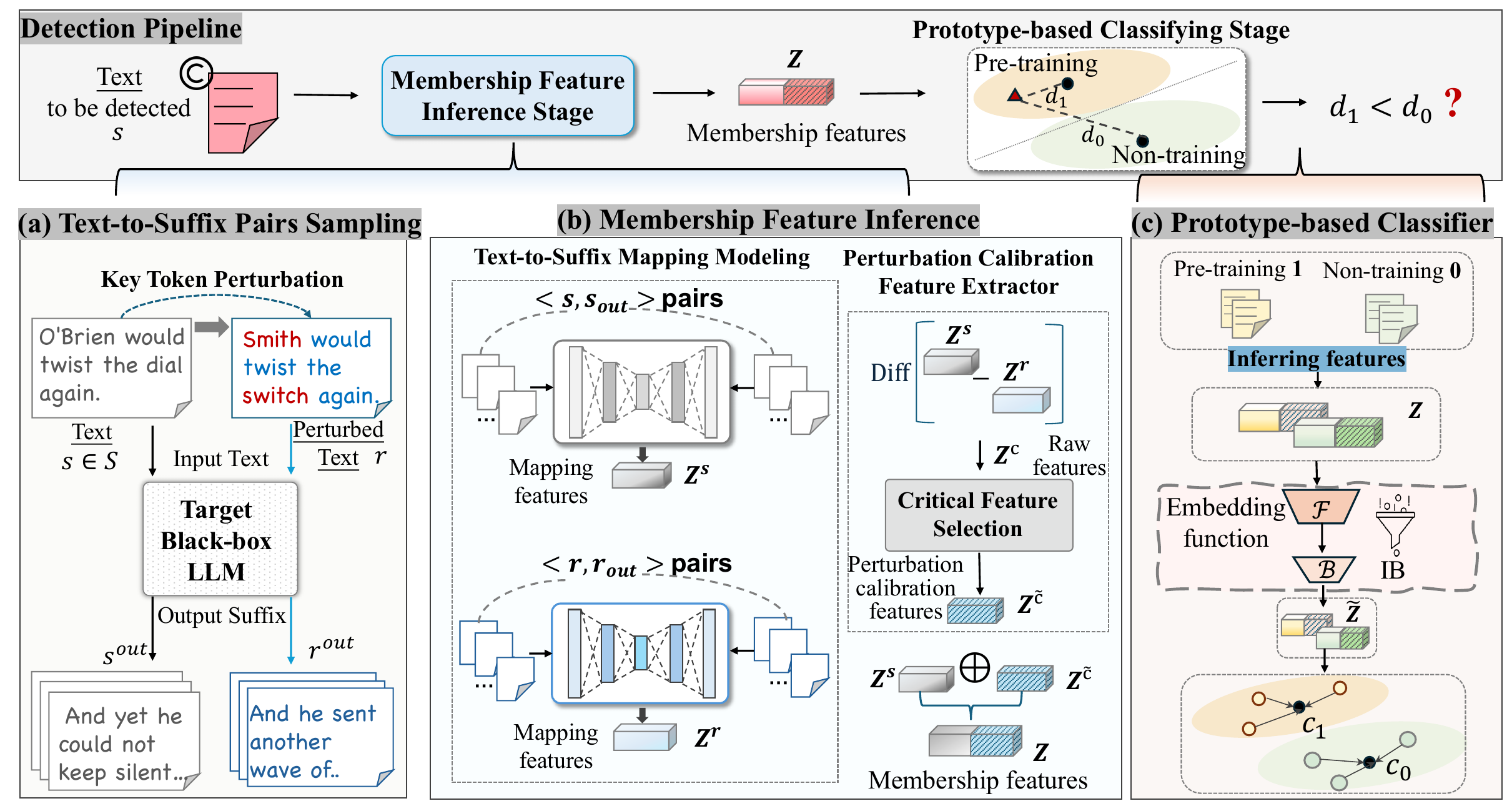}
  \caption{Overview of \textbf{VeilProbe}. 
 (a) The text-to-suffix sampling module generates text-to-suffix pairs; (b) The membership feature inference module infers membership features with a sequence-to-sequence model based on the above pairs; (c) The prototype-based classifier is trained based on the features from ground-truth samples, and then the pre-training and non-training prototypes are constructed. The first two modules prepare the membership feature inference stage for the texts to be detected, and the third one trains a membership classifier for detection.}
  \label{fig: method}
\end{figure*}

The purpose of this study is to detect texts in the black-box setting without human involvement. One natural idea is to infer the membership distinguishable features and use these features to train a membership classifier to identify pre-training texts, as shown in Fig.~\ref{fig: intro} (c). However, it is non-trivial due to the following two challenges. (1) The commonly used intuitive indicators~\cite{dong2024generalizationmemorizationdatacontamination,abbasi-yadkori2024to}, which are directly computed based on the input and output texts, such as semantic shifts or consistency of output texts, have been proven to be inadequate to distinguish members (see our experiments in Appendix~\ref{app:intuitive}); (2) The number of ground-truth pre-training samples is quite limited as the LLM providers usually do not disclose~\cite{karamolegkou2023copyrightviolationslargelanguage,shi2024detecting,chang2023speakmemoryarchaeologybooks}. This would result in overfitting issues when training the membership classifier.

To tackle the above challenges, we propose a novel pre-training text detection framework \textbf{VeilProbe}, consisting of two main parts: the membership feature inference module, and the prototype-based classifier. To capture the distinguishable features between training and non-training texts, the inference module first uses a sequence-to-sequence mapping model to simulate the mapping pattern of how the LLM completes a given text to its suffix. Additionally, inspired by ~\cite{10646875}, it integrates the perturbation calibration features by perturbing the key tokens in the input text to enrich the distinguishable features. Using these features, a prototype-based classifier~\cite{snell2017prototypicalnetworksfewshotlearning} integrated with Information Bottleneck (IB)~\cite{tishby2015deeplearninginformationbottleneck} is proposed for the detection of pre-training texts. This classifier is capable of filtering out irrelevant features. Moreover, it allows for effective classification using only a limited number of ground-truth training instances.
Our contributions are as follows:
\begin{itemize}
   \item We propose a novel framework named \textbf{VeilProbe}. To the best of our knowledge, we are the first to automatically detect pre-training text in the black-box LLMs without human involvement. 

    \item We propose a sequence-to-sequence model aiming to capture the intricate Text-to-Suffix pattern and infer the features that can distinguish membership. Additionally, we design a prototype-based classifier, which is capable of identifying the membership training data by leveraging merely a limited number of known samples.

    \item We compare the performance of VeilProbe to a wide range of baselines. It shows state-of-art results on three widely used datasets and outperforms the existing methods by a good margin. 
\end{itemize}

\section{Related Work}

\subsection{Membership Inference Attack in LLMs}
Membership Inference Attack (MIA)~\cite{carlini2022membership,Shokri_Stronati_Song_Shmatikov_2017} aims to identify whether a specific data point is in the model's training set. The concept of MIA, which was initially applied in the Computer Vision domain~\cite{Shokri_Stronati_Song_Shmatikov_2017}, has recently been extended to the Natural Language Processing tasks~\cite{carlini2021extracting,Ye_Maddi_Murakonda_Bindschaedler_Shokri_2022,duan2024membership}. MIA on LLMs can serve as a technique for several relevant tasks, such as Data Extraction Attacks~\cite{carlini2021extracting,yu2023bagtrickstrainingdata} and Personally Identifiable Information Attacks~\cite{lukas2023analyzingleakagepersonallyidentifiable,shi2024detecting,kim2023propileprobingprivacyleakage}, data contamination~\cite{oren2023provingtestsetcontamination,golchin2024timetravelllmstracing} and finetuning data detection~\cite{Mireshghallah_Goyal_Uniyal_Berg-Kirkpatrick_Shokri,Song_Shmatikov_2019}. Our works focus on an instance of MIA, pre-training data detection,  a task that has become a hot issue in recent research.

\subsection{Pre-training Data Detection}
%
According to the access level for detectors, there are three categories: the white-box, grey-box, and black-box settings. As the settings become increasingly strict, the information available to the detector diminishes, making the detection more challenging. Most studies focus on the grey-box setting, where the token probability distribution is accessible. Most of them~\cite{carlini2021extracting,mattern-etal-2023-membership,shi2024detecting,zhang2024mink,zhang2024pretrainingdatadetectionlarge} typically rely on heuristic assumptions to calculate a membership score for identifying pre-training texts.  Recently, some studies~\cite{maini2024llmdatasetinferencedid,puerto2024scalingmembershipinferenceattacks}selectively combine the membership scores as the aggregated feature.
We focus on the more challenging black-box setting. Some previous works~\cite{duarte2024decopdetectingcopyrightedcontent,chang2023speakmemoryarchaeologybooks}in such settings usually design complicated tasks to identify pre-training texts. Unlike previous work, our work focuses on automating the detection process in the black-box setting without human intervention.
 
\section{Problem Statement}
Formally, given a text $s\in S$, where the $S$ denotes the set of texts to be detected, and a target LLM $\Theta$ with pre-training data $ D_{\theta}$, we aim to determine whether $s$ is a member in the pre-training dataset $D_{\theta}$ of $\Theta$. Building upon previous work~\cite{shi2024detecting,zhang2024pretrainingdatadetectionlarge}, the task is defined as follows:
\[
\mathcal{A}(s| \Theta, \mathcal{K}) \rightarrow \{0, 1\}, \tag{1}
\]
where $\mathcal{A}$ denotes the detection algorithm, and $\mathcal{K}$ represents the information available to the detector. In the black-box setting, the detector can only access the input and output texts generated by $\Theta$.
We assume a few ground-truth pre-training and non-training text samples are available, which is in line with the practical scenarios. In particular, for the pre-training samples, a few classic popular published books, papers, and Wikipedia pages are well known to be included in the pre-training corpus~\cite{chang2023speakmemoryarchaeologybooks,karamolegkou2023copyrightviolationslargelanguage,7410368}. Moreover, there are also some unpublished manuscripts from some writers that can be the non-training samples, which is consistent with the setting in previous studies such as~\cite{maini2024llmdatasetinferencedid,puerto2024scalingmembershipinferenceattacks}.

\section{Method}
Our proposed \textbf{VeilProbe} framework is illustrated in Figure~\ref{fig: method}, which is composed of three modules: (1) For preprocessing, the text-to-suffix pairs sampling module is proposed to sample the completion suffix $s^{\text{out}}$ of each input text $s \in S$ by requesting the target LLM response. Moreover, we perturb the key tokens of $s$ to generate $r$, which is then fed into LLMs to obtain the corresponding suffix $r^{\text{out}}$. (2) Based on multiple $\langle s, s^{\text{out}}\rangle$ and  $\langle r, r^{\text{out}}\rangle$ pairs, the membership feature inference module is devised to infer the latent membership feature $\bm{Z}$. (3) The prototype-based classifier is trained with the inferred membership features to obtain the prototypes for pre-training and non-training data.
In the detection pipeline, given a text $s$ to be detected, if its membership feature is closer to the pre-training prototype, it can be inferred that $s$ is a member of the target LLM's pre-training data.


\subsection{Text-to-Suffix Pairs Sampling}\label{sec: sampling}
As shown in Figure~\ref{fig: method} (a), for preprocessing, the sampling stage aims to generate a set of text-to-suffix pairs from the target black-box LLM, including text-to-suffix pairs $\langle s, s^{\text{out}} \rangle$ and perturbed text-to-suffix pairs  $\langle r, r^{\text{out}} \rangle$. 

\paragraph{Text-to-Suffix Pair Generation.}
The autoregressive completion task is employed to align with the training patterns of the corpus during the pre-training phase. To be specific, without additional instructions, each text $s$ serves as a prefix, and the target black-box LLM $\Theta$ is requested to complete the sentence based on $s$. The target LLM autoregressively generates the subsequent tokens until the maximum generation length is reached.
A pair of the text $s$ and its corresponding response $s^{\text{out}}$ is denoted as $\langle s, s^{\text{out}} \rangle$. More than one suffix is generated for each text to reduce the randomness in the responses. Similarly, the corresponding suffix $r^{\text{out}}$ of the perturbed text $r$  is generated in the same autoregressive way. The following will elaborate on how to obtain the perturbed \(r\).


\paragraph{Key Token Perturbation.}
We carry out perturbations on the key tokens within the original input $s$ to generate $r$. The key tokens are those that exert substantial influence on the generation of the suffix text.
Inspired by ~\cite{ligilot}, we employ a feature attribution technique to identify key tokens. However, directly applying feature attribution analysis to the black-box LLM is non-trivial.  Most of these methods depend on hidden information (prediction or token probability distributions), which is unsuitable for the black-box scenario.
 
\begin{figure}[t]
  \includegraphics[width=1\columnwidth]{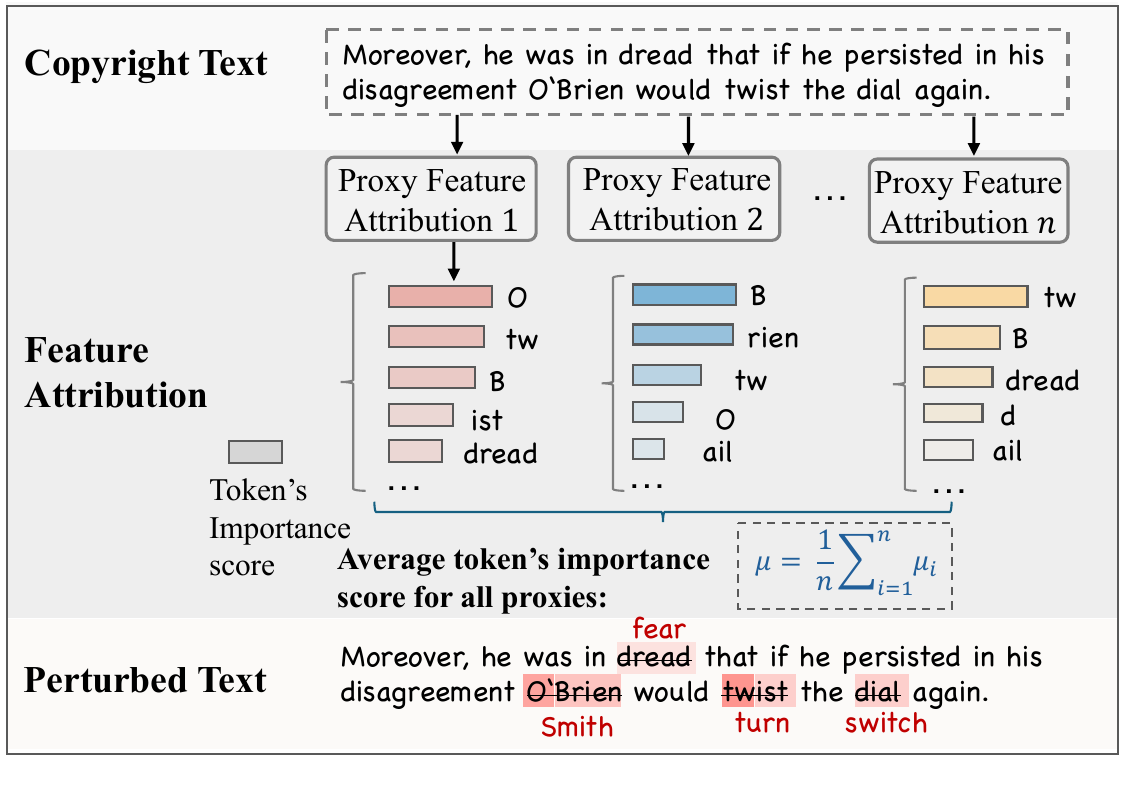}
  \caption{The pipeline of key token perturbation module.}
  \label{fig: keyperturb}
\end{figure}

To address the above issue, a perturbation-based feature attribution method is introduced, as illustrated in Figure~\ref{fig: keyperturb}.  We use multiple small-size open-source LLMs (preferably from the same family) as proxy models to assist in selecting the key tokens. It is observed that when using LLMs from the same family as proxy LLMs, the feature attribution technique tends to select a similar set of informative tokens in a text. For instance, GPT-2 can serve as a proxy LLM for ChatGPT. (See Appendix~\ref{app: keytoken} for detailed analysis). Then the importance scores obtained for each token across multiple proxy LLMs are averaged to obtain the final importance score $\mu$. Based on these scores, top-$\gamma\%$ tokens are selected and perturbed with synonyms to produce the perturbed text $r$. 
 

\subsection{Membership Feature Inference}
This module is designed to infer membership features based on the $\langle s, s^{\text{out}} \rangle$ and $\langle r, r^{\text{out}} \rangle$ pairs obtained from the above module.  

\paragraph{Text-to-Suffix Mapping Modeling.}
Since LLMs generate outputs one by one in an autoregressive manner, the input-output mapping $f$ generally follows a sequence-to-sequence pattern~\cite{sutskever2014sequencesequencelearningneural,vaswani2023attentionneed}. To capture such a pattern, a sequence-to-sequence mapping model based on the transformer architecture~\cite{vaswani2023attentionneed} is proposed to model the mapping pattern from the input text to its suffix. Specifically, this mapping model is trained on $\langle s, s^{\text{out}} \rangle$ pairs to learn the input-output dependency, that is, $f: s \rightarrow s^{\text{out}}$. The training objective is to maximize the conditional probability of the output sequence $s^{\text{out}}$ given the input sequence $s$ as the prefix, which is expressed as:
\[
P(s^{\text{out}} |s) = \prod_{t=1}^{T} P(w_t^{\text{out}} | w_1^{\text{out}}, w_2^{\text{out}}, \ldots, w_{t-1}^{\text{out}}, s)\tag{2},
\]
where $T$ denotes the max length of the output sequence $s^{\text{out}}$, and each $w_t^{\text{out}}$ represents the token generated at time step $t$. The model learns to predict $w_t^{\text{out}}$ based on the previously generated tokens $w_1^{\text{out}}, \ldots, w_{t-1}^{\text{out}}$ and the input sequence ${s}$.  
Upon completion of the training process, once a text input is provided to the sequence-to-sequence model, the hidden states of the model are extracted as the text-to-suffix mapping feature $\bm{Z^{\text{s}}}$. Similarly, the perturbed mapping features $\bm{Z^{\text{r}}}$ can be obtained based on $\langle r, r^{\text{out}} \rangle$ pairs.

The obtained mapping feature can function as a part of the membership distinguishable features to distinguish between pre-training and non-training texts. The fundamental concept is based on previous research~\cite{wang2024pandoraswhiteboxprecisetraining} which has shown that the hidden states in the internal layers of a white-box LLM can tell the difference between the pre-training texts and non-training texts. 

\paragraph{Perturbation Calibration Feature Extractor.}
In addition to the hidden states obtained from the aforementioned mapping model, we further explore the perturbation calibration feature as a complement.  The intuition is based on the observation that the robustness of the target model exhibits marked differences when perturbed pre-training samples and non-training samples are employed as inputs~\cite{10646875,mattern-etal-2023-membership}.

Specifically, as shown in Figure~\ref{fig: method} (b), we first obtain the raw perturbation feature $\bm{Z^{\text{c}}}$, which is the difference between the original $\bm{Z^{\text{s}}}$ and perturbed mapping features $\bm{Z^{\text{r}}}$. 
Since the perturbation operations would probably induce noisy information~\cite{10646875,hooker2019benchmarkinterpretabilitymethodsdeep} for identifying members, it is crucial to select the most critical features to infer membership status. The significance test~\cite{10.1093/biomet/34.1-2.28} is applied to filter critical features. We formulate the null hypothesis ($H_0$) and the alternative hypothesis ($H_1$) for the values in each dimension of the feature with respect to their impact on the target variable (i.e., membership status $Y$). The hypotheses are defined as follows: 
\begin{align*}
H_{0i} &: \Pr(Y = 1 \mid z_i^{\text{c}} \in G_1) = \Pr(Y = 1 \mid  z_i^{\text{c}}\in G_0), \tag{3}\\
H_{1i} &: \Pr(Y = 1 \mid z_i^{\text{c}} \in G_1) \neq \Pr(Y = 1 \mid z_i^{\text{c}}\in G_0), \tag{4}
\end{align*}
where $G_1$, and $G_0$ represent the ground-truth pre-training and non-training samples, respectively. $z_i^{\text{c}}$ represents the $i$-th dimension of the feature vector $\bm{Z^{\text{c}}}$. Then the $p$-value of each dimension of the feature vector is obtained. A lower $p$-value indicates strong evidence against the null hypothesis, suggesting that the pre-training and non-training samples are distinct in the $i$-th dimension, supporting the alternative hypothesis. We then retain the critical dimensions and set the others to zero. The processed features then serve as the final perturbation calibration feature $\bm{Z^{\widetilde{\text{c}}}}$.

Consequently, the membership feature of $s$ is the concatenation of the mapping feature $\bm{Z^{\text{s}}}$ and the perturbation calibration feature $\bm{Z^{\widetilde{\text{c}}}}$, which is defined as:
\[
  \bm{Z} = \bm{Z^{\text{s}}} \oplus\bm{Z^{\widetilde{\text{c}}}}. \tag{5}
\]

\subsection{Prototype-based Classifier}
Based on the membership features, we propose a prototype-based classifier to distinguish between pre-training and non-training texts. We choose this classifier as a backbone because it works well even with limited labeled data.


\paragraph{Prototype  Constructing.} The prototype-based classifier constructs a metric space through learning~\cite{snell2017prototypicalnetworksfewshotlearning}. In this space, samples belonging to the same class are proximal to each other, whereas samples from different classes are distantly separated.

Each class is represented by a prototype, which is the center of its samples in this space. 
As shown in Figure~\ref{fig: method} (c), based on the membership features, we aim to construct the pre-training prototype and the non-training prototype. Specifically, the embedding function $\mathcal{F}$ is trained following the approach of Prototypical Networks~\cite{snell2017prototypicalnetworksfewshotlearning} to mitigate overfitting issues caused by the limited labeled data. The training episodes are constructed by randomly sampling a subset of samples from the ground-truth set $G$ to form the support set $U$, while a separate subset of the remaining samples serves as query points. 
 
Formally, we define the support set as $U=\{(\bm{Z}_1, Y_1),\ldots, (\bm{Z}_m, Y_m)\}$, where each $\bm{Z}_i \in \mathbb{R}^D$ represents the $D$-dimensional feature vector of an sample, and $Y_i \in \{0, 1\}$ is the corresponding label. $U_Y$ represents the set of samples labeled with pre-training ($U_1$) or non-training ($U_0$).
The objective is to compute an $K$-dimensional representation, denoted as $c_Y \in \mathbb{R}^K$, where $c_0$ represents the non-training prototype and  $c_1$ represents the training prototype. These prototypes can be obtained using an embedding function $\mathcal{F}_\phi: \mathbb{R}^D \to \mathbb{R}^K$ with learnable parameters $\phi$.
Each prototype is defined as the mean vector of the embedded representations of the support samples belonging to their set:

\[
c_Y = \frac{1}{|U_Y|} \sum_{(\bm{Z}_i, Y_i) \in U_Y} \mathcal{F}_\phi(\bm{Z}_i) \tag{6}.
\]
We define the probability distribution over class labels for a query point \(\bm{Z}\) using a softmax function, which is based on the distances between the query's embedding and the class prototypes. The probability of classifying \(\bm{Z}\) as belonging to class \(Y\) is given by:

\[
p_\phi(y = Y \mid \bm{Z}) = \frac{\exp(-d(\mathcal{F}_\phi(\bm{Z}), c_Y))}{\sum_{Y'} \exp(-d(\mathcal{F}_\phi(\bm{Z}), c_{Y'}))}, \tag{7}
\]
where \(d(\mathcal{F}_\phi(\bm{Z}), c_Y)\) represents the distance between the query embedding \(\mathcal{F}_\phi(\bm{Z})\) and the prototype \(c_Y\) of class \(Y\), and the distance function \(d\) is squared Euclidean distance. The learning process minimizes the negative log-likelihood of the true class label via Stochastic Gradient Descent.

\paragraph{Information Bottleneck Denoising.}
To further filter out redundant features in $\bm{Z}$, we integrate the Information Bottleneck technique (IB) into the prototype computation. 
This technique aims to extract a compact representation that retains the information relevant to $Y$ while eliminating redundant information from the input feature $\bm{Z}$.
Specifically, we construct a compressed representation $\bm{\tilde{Z}}$ by optimizing the following objective:
\[
\min \mathcal{L}_{\text{IB}} = I(\bm{Z},\bm{\tilde{Z}}) - \beta I(\bm{\tilde{Z}}; Y),\tag{8}
\]
where $I(\cdot)$ represents the mutual information, $\bm{Z}$ represents the input feature, $Y$ is the ground-truth label, and $\beta$ balances the trade-off between minimizing redundancy and retaining relevance to $Y$. 

We then use the compressed representation $\bm{\tilde{Z}}$ as the feature to detect the pre-training text $s$. That is, the detection score $\delta$ is defined as: 
\[
\delta = d(\bm{\tilde{Z}}, c_0) - d(\bm{\tilde{Z}}, c_1) ,\tag{9}
\]
where $d$ represents 
the distance metric, $c_{1}$ represents the pre-training prototype, and $c_{0}$ represents the non-training prototype.
If $\bm{\tilde{Z}}$ is closer to the $c_{1}$ than $c_{0}$, it can be inferred that the text $s$ is a member in the pre-training data of the target LLM.

\section{Experiments}
\subsection{Experimental settings}\label{sec: settings}
\noindent\textbf{Datasets}. The statistics of three datasets are shown in Table~\ref{tab: dataset}. \textbf{WikiMIA}~\cite{shi2024detecting} consists of Wikipedia event snippets. \textbf{BookTection}~\cite{duarte2024decopdetectingcopyrightedcontent} is a widely adopted dataset that contains 165 copyrighted books, which is expanded based on BookMIA \cite{shi2024detecting}. \textbf{arXivTection}~\cite{duarte2024decopdetectingcopyrightedcontent} includes classic papers from arXiv. 


\begin{table}[t]
    \centering
    \huge
    \renewcommand{\arraystretch}{0.8} 
    \setlength{\belowcaptionskip}{-2pt}
    \resizebox{1\columnwidth}{!}{%
    \begin{tabular}{lccc}
        \toprule
        \textbf{Datasets}& \textbf{Len.} & \textbf{\#Samples}& \textbf{Resources} \\
        \midrule
       \multirow{4}{*}{WikiMIA~\cite{shi2024detecting} }& 32 & 776 & \multirow{4}{*}{Wikipedia events} \\
         & 64 & 542 & \\
         & 128 & 250 & \\
         & 256 & 82 &\\
       \midrule
        \multirow{3}{*}{BookTection~\cite{duarte2024decopdetectingcopyrightedcontent}}
         & 64 & 5472 &\multirow{3}{*}{165 Bestselling books} \\ 
         & 128 & 5494 & \\
         & 256 & 5447 & \\
          \midrule
        \multirow{1}{*}{arXivTection~\cite{duarte2024decopdetectingcopyrightedcontent}} 
         & 128 & 1549 & 50 arXiv papers\\
        \bottomrule
    \end{tabular}}
    \caption{The statistics of three datasets. }
    \label{tab: dataset}
\end{table}

\noindent\textbf{Target LLMs}. Following~\cite{shi2024detecting}, for WikiMIA, we select Pythia-6.9B~\cite{biderman2023pythiasuiteanalyzinglarge}, LLaMA-13B~\cite{touvron2023llamaopenefficientfoundation}, and GPT-NeoX-20B~\cite{black2022gptneox20bopensourceautoregressivelanguage} as our target LLMs. Following~\cite{duarte2024decopdetectingcopyrightedcontent}, for BookTection, Mistral-7B~\cite{jiang2023mistral7b}, LLaMA2-7B~\cite{touvron2023llama2openfoundation} and  ChatGPT (gpt-3.5-instruct) are selected as the target LLMs. For arXivTection, Mistral-7B~\cite{jiang2023mistral7b}, LLaMA2-13B~\cite{touvron2023llama2openfoundation}, and Claude 2.1 are selected as our target LLMs.

\noindent\textbf{Evaluation Metrics}.
Following previous works~\cite{shi2024detecting,zhang2024pretrainingdatadetectionlarge,duarte2024decopdetectingcopyrightedcontent,carlini2022membership} we use the Area Under the ROC curve (AUC) and True Positive Rate at low False Positive Rate (TPR@5\%FPR) as our evaluation metrics. A higher AUC reflects better performance with higher TPR and lower FPR across thresholds.

\begin{table*}[htbp]
    \centering
    \huge
    \renewcommand{\arraystretch}{0.9} 
    \resizebox{1\textwidth}{!}{
    \begin{tabular}{ll|ccc|ccc|ccc}
        \specialrule{1.4pt}{0pt}{0pt}
        \textbf{Type} & \textbf{Method} & \multicolumn{3}{c}{\textbf{WikiMIA}} & \multicolumn{3}{c}{\textbf{BookTection}}&\multicolumn{3}{c}{\textbf{arXivTection}} \\
        \cmidrule(lr){3-5}\cmidrule(lr){6-8}\cmidrule(lr){9-11}
        & & Pythia-6.9B & LLaMA-13B & NeoX-20B & Mistral-7B & LLaMA2-7B & ChatGPT &Mistral-7B& LLaMA2-13B & Claude 2.1\\
        \midrule
        \multirow{7}{*}{Grey-box} 
        & PPL            & 0.635 &0.666 & 0.689 & 0.698 & 0.701  & \ding{55} &0.684& 0.658 & \ding{55} \\
        & Lowercase      & 0.600 & 0.602 &0.666 & 0.675  &0.697 & \ding{55} &0.547&0.475& \ding{55} \\
        & Zlib           & 0.645&  0.678  & 0.700  & 0.539  &0.541 & \ding{55}&0.586& 0.566 & \ding{55} \\
        & Neighbor       & 0.649 &  0.654 & 0.690 &0.614 &0.637& \ding{55}&0.525& 0.528  & \ding{55} \\
        & Min-K\% Prob   & 0.663& 0.688& 0.736  &0.698  & 0.701 & \ding{55}&0.749 & 0.732  & \ding{55} \\
        & Min-K\%++ Prob &0.691& 0.847 & 0.755 &0.593 & 0.610  & \ding{55}&0.695&0.708 & \ding{55} \\
        & DC-PDD         & 0.698 &  0.697 & 0.766 & --  &  -- & \ding{55}&-- & --    & \ding{55} \\
        & FeartureAgg  & 0.651 & 0.645 &0.728& 0.696 &0.702   & \ding{55}&0.774& 0.751  & \ding{55} \\
        \midrule
        \multirow{2}{*}{Black-box} 
          & Name-cloze         & 0.557 &0.534 & 0.525 &  0.538 &0.538& 0.544 & 0.540 & 0.527 & 0.609  \\
        & DE-COP         &0.512 &0.512 & 0.531&  0.674  &0.616& 0.720 &0.551 & 0.506   &0.583\\
        & \textbf{VeilProbe} & \bf{0.913} & \bf{0.902} & \bf{0.916}   & \bf{0.960}& \bf{0.963}&\bf{0.947}&\bf{0.797} & \bf{0.837} & \bf{0.904} \\
       \specialrule{1.4pt}{0pt}{0pt}
    \end{tabular}%
    }
    \setlength{\abovecaptionskip}{7pt}
    \caption{AUC scores on WikiMIA, BookTection-128, and arXivTection with corresponding target LLMs.}
    \label{tab: mainauc}
\end{table*}

\begin{table*}[htbp]
    \centering
    \huge
     \renewcommand{\arraystretch}{0.9} 
    \resizebox{1\textwidth}{!}{
    \begin{tabular}{ll|ccc|ccc|ccc}
      \specialrule{1.4pt}{0pt}{0pt}
   \textbf{Type} & \textbf{Method} & \multicolumn{3}{c}{\textbf{WikiMIA}} & \multicolumn{3}{c}{\textbf{BookTection}}&\multicolumn{3}{c}{\textbf{arXivTection}} \\
        \cmidrule(lr){3-5}\cmidrule(lr){6-8}\cmidrule(lr){9-11}
        & & Pythia-6.9B & LLaMA-13B & NeoX-20B & Mistral-7B & LLaMA2-7B & ChatGPT  &Mistral-7B& LLaMA2-13B & Claude 2.1\\
        \midrule
        \multirow{7}{*}{Grey-box}
        & PPL            & 0.140 & 0.156  & 0.170 & 0.214  &0.220& \ding{55} & 0.095&0.083  & \ding{55} \\
        & Lowercase      &0.117 & 0.123 & 0.152 &0.209 &0.243& \ding{55} &0.087&  0.058 & \ding{55} \\
        & Zlib           &  0.178  & 0.143 & 0.196 & 0.134 & 0.150& \ding{55} &0.067&  0.062 & \ding{55} \\
        & Neighbor       & 0.045   & 0.116   & 0.170   & 0.132  & 0.144 & \ding{55} &0.045& 0.049 & \ding{55} \\
        & Min-K\% Prob   &  0.183  & 0.187  & 0.233 & 0.214 & 0.220& \ding{55} &0.197& 0.166 & \ding{55} \\
        & Min-K\%++ Prob &  0.211  &  0.369  & 0.214 &0.118 &0.148 & \ding{55} &0.143& 0.162 & \ding{55} \\
        & DC-PDD         & 0.245  &  0.230 &  0.317 & --    & --    & \ding{55} & --    &--& \ding{55} \\
        & FeartureAgg & 0.166 & 0.120& 0.229  &0.238& 0.218 & \ding{55}& 0.257 &0.241 & \ding{55} \\
        \midrule
        \multirow{2}{*}{Black-box}
           & Name-cloze         & 0.093  &0.070& 0.079  & 0.127 &0.114& 0.148 & 0.124  &  0.096   & 0.090 \\
        & DE-COP         &0.061    &  0.064   &0.092&0.208 & 0.173   & 0.385&0.067& 0.018  & 0.079\\
        & \textbf{VeilProbe} & \bf{0.625} & \bf{0.554}    & \bf{0.625}   &\bf{0.807} & \bf{0.845}    & \bf{0.723} &\bf{0.367}&\bf{0.411}  & \bf{0.581} \\
       \specialrule{1.4pt}{0pt}{0pt}
    \end{tabular}%
    }
       \setlength{\abovecaptionskip}{7pt}
    \caption{TPR@5\%FPR scores on WikiMIA, BookTection-128, and arXivTection with corresponding target LLMs.}
    \label{tab: maintprfpr}
\end{table*}

\begin{table}[t]
    \centering
    \huge
    \setlength{\belowcaptionskip}{-1pt}
   \renewcommand{\arraystretch}{0.85} 
    \resizebox{1\columnwidth}{!}{%
    \begin{tabular}{l|l|ccc|c}
        \specialrule{1.25pt}{0pt}{0pt}
        \textbf{Document} & \textbf{Metric} & \textbf{Mistral-7B} &\textbf{LLaMA2-13B} & \textbf{Claude 2.1}& \textbf{Avg.} \\
        \cmidrule[0.8pt](lr){3-6}
        \multirow{2}{*}{165 books} & AUC & 0.996 & 0.998 & 0.988& 0.994\\
        & TPR@5\%FPR & 0.991&1.000 & 0.943& 0.978  \\
        \hline\hline
        &\textbf{Metric} &\textbf{Mistral-7B}&\textbf{LLaMA2-7B}&\textbf{ChatGPT} &\textbf{Avg.} \\
         \cmidrule[0.8pt](lr){3-6}
          \multirow{2}{*}{50 papers}& AUC &0.978& 0.994 & 0.994 & 0.989 \\
        &  TPR@5\%FPR  & 0.920 & 0.960& 0.960 &0.947
       \\
        \specialrule{1.25pt}{0pt}{0pt}
    \end{tabular}}
    \caption{The detection performance of 165 books and 50 papers.}
    \label{tab: document}
\end{table}

\noindent\textbf{Baselines}. We compare our framework with the following strong baselines. For the grey-box methods, eight methods are selected: \textbf{PPL}, \textbf{Lowercase}, \textbf{Zlib}~\cite{carlini2021extracting}, \textbf{Neighbor}~\cite{mattern-etal-2023-membership}, \textbf{Min-K\% Prob}~\cite{shi2024detecting}, \textbf{Min-K\%++ Prob}~\cite{zhang2024mink}, \textbf{DC-PDD}~\cite{zhang2024pretrainingdatadetectionlarge}, and \textbf{FeatureAgg}~\cite{maini2024llmdatasetinferencedid}. Two methods are selected for the black-box methods: \textbf{DE-COP}~\cite{duarte2024decopdetectingcopyrightedcontent} and \textbf{Name-Cloze}~\cite{chang2023speakmemoryarchaeologybooks}. The details are described in Appendix~\ref{app: baseline}.

\noindent\textbf{Implementation Details}. In our work, all experiments are implemented on a workstation with five NVIDIA Tesla V100 32G GPUs, and Ubuntu22.04.4. For each text to be detected, three suffixes were generated using the target LLM, with the maximum suffix length set to 512 tokens. The parameter $\gamma$ is set to 10 for obtaining the perturbed text $r$. The $p$-value significance threshold is chosen from  \{0.001, 0.01, 0.05, 0.1\} to select the critical perturbation calibration features. We randomly sample approximately 50 ground-truth samples per dataset to train the prototype-based classifier, with the remaining samples serving as the texts to be detected.
It should be noted that our evaluations are carried out at both sentence and document levels, and the document-level results are derived from the sentence-level ones. Specifically, we compute the average of the sentence-level detection scores within each document to obtain the document-level results.

\subsection{Main Results}\label{sec:mainresults}
 Table~\ref{tab: mainauc} and Table~\ref{tab: maintprfpr} show the comparison methods' AUC and TPR@5\%FPR scores, respectively. We can observe that (1) VeilProbe significantly outperforms all the other methods, achieving an impressive improvement of over 25.1\% in AUC and nearly doubling the TPR@5\%FPR scores; (2) Compared to grey-box detection methods, VeilProbe can be implemented on ChatGPT and Claude because it does not rely on hidden information; (3) VeilProbe significantly improves text detection performance compared to existing black-box methods. One possible reason is that these methods often rely on complicated instructions, making some LLMs struggle to comprehend the task. 
 
Additionally, we evaluate document-level detection on 165 bestselling books and 50 arXiv papers from the BookTection and arXivTection datasets by aggregating sequences' detection scores. As shown in Table~\ref{tab: document}, VeilProbe achieves an average AUC of 99.2\% and an average TPR@5\%FPR of 96.3\% when detecting documents. The detailed results are provided in the Appendix~\ref{app: book}. The document-level results consistently outperform the sentence-level detection, which is in line with previous works~\cite{puerto2024scalingmembershipinferenceattacks}. 
\begin{table}[t]
\centering
\huge
 \renewcommand{\arraystretch}{0.8} 

\setlength{\belowcaptionskip}{-1pt}
\resizebox{0.9\columnwidth}{!}{%
\begin{tabular}{@{}l|c|ccc@{}}
\specialrule{1.25pt}{0pt}{0pt}
\textbf{Metric} & \textbf{Approaches} & \textbf{WikiMIA} & \textbf{BookTection} \\ \midrule
\multirow{7}{*}{AUC} 
  & w/o calibration feature & 0.929 & 0.912 \\
  & w/o critical selection &0.935& 0.908 \\
  & w/o key perturbation &0.931& 0.914 \\
   \cmidrule[0.1pt](lr){2-4}
  & w/o noise reduction & 0.929 & 0.915 \\
  & Cosine distance & 0.887 & 0.904 \\ 
  & Manhattan distance & 0.934 & 0.917 \\ 
  \cmidrule[0.1pt](lr){2-4}
  & Full  & \bf{0.937} & \bf{0.919} \\ 
\midrule
\multirow{7}{*}{TPR@5\%FPR} 
  & w/o calibration feature & 0.668& 0.622 \\
  & w/o critical selection & 0.684& 0.625 \\
  & w/o key perturbation & 0.703& 0.637 \\
   \cmidrule[0.1pt](lr){2-4}
  & w/o noise reduction & 0.719& 0.631 \\
  & Cosine distance & 0.476 &  0.649 \\ 
  & Manhattan distance & 0.668 & \bf{0.653} \\ 
  \cmidrule[0.1pt](lr){2-4}
  & Full & \bf{0.731} & 0.644 \\ 
 \specialrule{1.25pt}{0pt}{0pt}
\end{tabular}}
\caption{Ablation studies on the WikiMIA with Pythia-6.9B and on the BookTection with ChatGPT as the corresponding target LLMs.}
\label{tab: ablation}
\end{table}

\subsection{Ablation Studies}
To further assess the importance of each module in \textbf{VeilProbe}, we perform ablation studies on WikiMIA-64 and BookTection-64. For the membership feature inference stage, we analyze the contributions of (i) perturbation calibration feature extraction, (ii) critical feature selection with significance tests, and (iii) key token perturbation. For the prototype-based classifier, we evaluate (i) the impact of the IB module and (ii) the impact of different distance metrics for prototype computation.

Table~\ref{tab: ablation} shows that all ablation variants demonstrate lower performance than the full framework. The perturbation calibration features are essential, particularly in the TPR@5\%FPR score. The critical feature selection can improve performance by reducing noise induced by perturbation operations. Key token perturbations are more effective than random perturbations. The IB module can eliminate redundant information regarding membership status to enhance performance. The most suitable distance function for prototype computing in our task is the Squared Euclidean distance.

\begin{figure}[t]
  \centering
   \hspace*{-0.04\linewidth} 
  \begin{subfigure}[b]{0.48\linewidth}
    \centering
    \includegraphics[width=\linewidth]{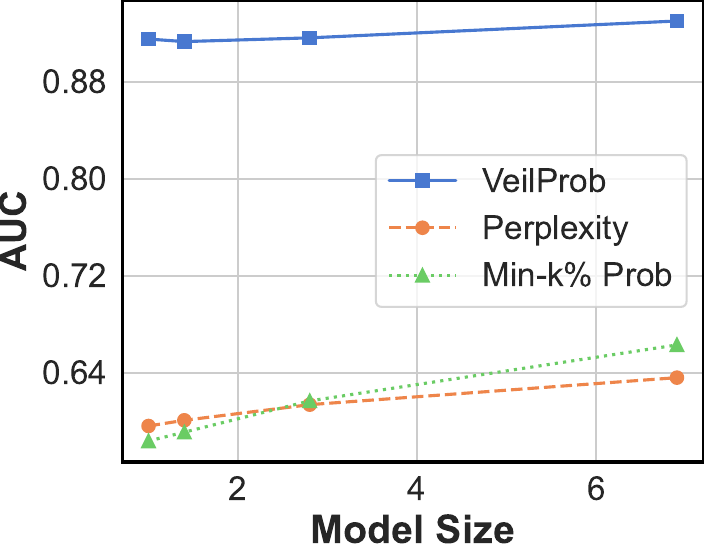}
    \caption{AUC w.r.t Model Size.}
    \label{fig:modelsizewiki1}
  \end{subfigure}
   \hspace{0.01\linewidth} 
  \begin{subfigure}[b]{0.48\linewidth}
    \centering
    \includegraphics[width=\linewidth]{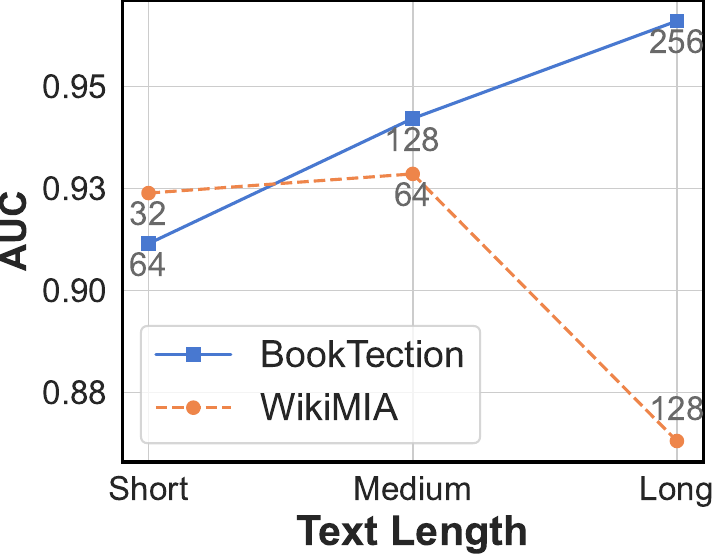}
    \caption{AUC w.r.t Text Length.}
    \label{fig:textlen}
  \end{subfigure}
  \caption{VeilProbe's performance w.r.t different general factors.}
  \label{fig:combined}
\end{figure}

\begin{figure}[t]
  \includegraphics[width=1\columnwidth]{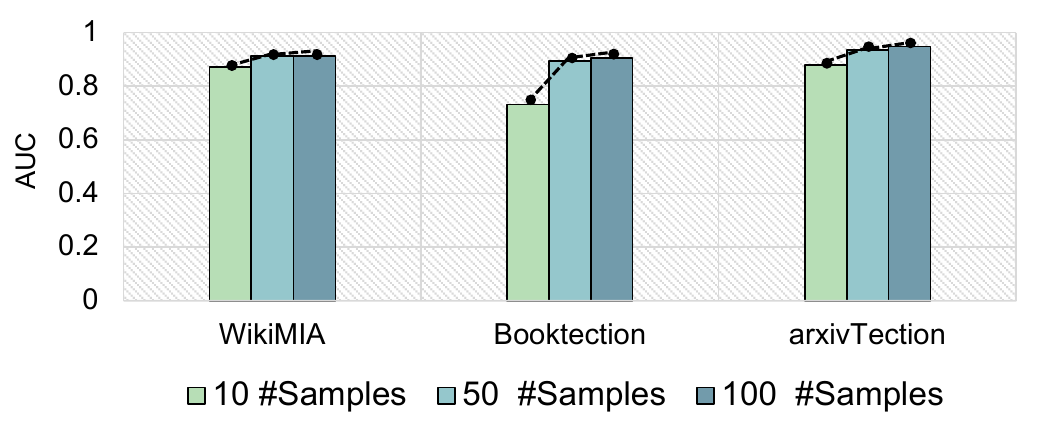}
  \caption{VeilProbe's performance w.r.t \#Ground-truth samples. }
  \label{fig: groundtruth}
\end{figure}

\subsection{Discussion and Analysis}
In this subsection, we analyze five main factors that may affect the detection performance. Among them, the former two are general factors, while the latter three are framework-specific factors. 
\paragraph{Size of target LLMs.} 
To investigate the impact of different target LLM sizes, we analyze Pythia models of sizes 1B, 1.4B, 2.8B, and 6.9B on WikiMIA. Figure~\ref{fig:modelsizewiki1} shows that AUC scores improve with increasing model sizes, aligning with previous studies~\cite{shi2024detecting,duarte2024decopdetectingcopyrightedcontent}. Additionally, our proposal consistently outperforms baselines.

\paragraph{Length of texts.} We evaluate performance with respect to different text lengths (Figure~\ref{fig:textlen}). For BookTection, the detection performance increases with longer texts. The possible reason is that longer texts can contain more distinguishable features. For WikiMIA, the AUC score decreases at the length of 128, potentially due to the limited number of texts of this length (only 250 samples), much smaller in number than the samples of other lengths, as shown in Table~\ref{tab: dataset}.

\paragraph{Number of ground-truth samples.} 
We further explore the impact of the number of ground-truth samples. Figure~\ref{fig: groundtruth} shows that AUC scores improve as the number of samples increases. Nevertheless, even with only 10 ground-truth samples, our method is still capable of achieving an AUC detection rate of approximately 80-90\% across three datasets.

\paragraph{Volume of text-to-suffix mapping features.} 
Table~\ref{tab: volume} shows the impact of different volumes of mapping features on the detection performance. 
We use four extraction methods: (i) \texttt{Full}, which extracts the hidden states from all layers of the mapping model; (ii) \texttt{Full-LT}, extracting the last token's hidden states from all layers; (iii) \texttt{LL}, extracting the hidden states from the last layer of both encoder and decoder; and (iv) \texttt{LLT}, extracting the last token's hidden states from the last layer of both encoder and decoder.
Despite a general slight decline in performance metrics as the quantity of mapping features is decreased, our approach stays effective at a high detection performance.

\begin{figure}[t]
  \centering
  \begin{subfigure}[b]{0.51\linewidth}
    \centering
    \includegraphics[width=\linewidth]{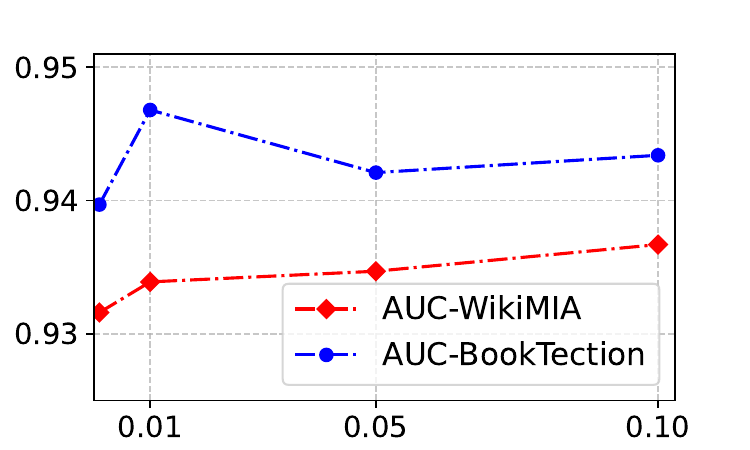}
    \caption{AUC.}
  \end{subfigure}
   \hspace{-0.04\linewidth} 
  \begin{subfigure}[b]{0.51\linewidth}
    \centering
    \includegraphics[width=\linewidth]{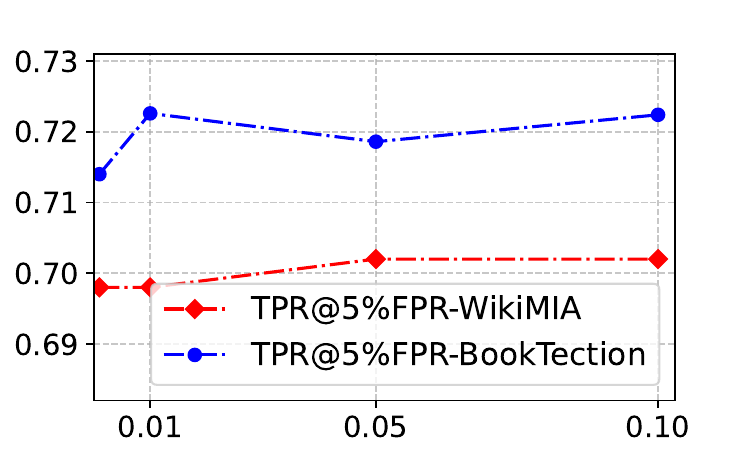}
    \caption{TPR@5\%FPR.}
  \end{subfigure}
  \caption{VeilProbe's performance w.r.t $p$-value.}
  \label{fig: pvalue}
\end{figure}

\paragraph{Significance threshold for critical perturbation calibration feature selection.} 
Figure~\ref{fig: pvalue} shows that the optimal $p$-value varies across different text domains.
A $p$-value of 0.1 allows tolerance, 0.05 is standard, and 0.001 is stricter. When the $p$-values are large, there is a risk of underrepresenting crucial features. Conversely, when they are small, it may lead to the introduction of misleading features that impede detection.

\begin{table}[t]
    \centering
    \huge
    \setlength{\belowcaptionskip}{-2pt}
     \renewcommand{\arraystretch}{0.8} 
    \resizebox{1\columnwidth}{!}{%
    \begin{tabular}{llcccccccc}
        \toprule
        \textbf{Dataset} & \textbf{Metric} & \texttt{Full} & \texttt{Full-LT} & \texttt{LL} & \texttt{LLT}& \textbf{Avg.} \\
        \midrule
        \multirow{2}{*}{WikiMIA} & AUC &\bf{0.937} & 0.928 & 0.924 & 0.916 & 0.926\\
        & TPR@5\%FPR& \bf{0.731} & 0.612 & 0.573 & 0.569 & 0.621 \\
        \midrule
        \multirow{2}{*}{BookTection} & AUC & \bf{0.947}& 0.937 & 0.939 & 0.928& 0.938 \\
        &  TPR@5\%FPR  &\bf{ 0.723} & 0.702& 0.689 & 0.663& 0.694\\
        \midrule
        \multirow{2}{*}{arXivTection} & AUC & 0.890 & 0.880 & \bf{0.904} & 0.890& 0.891 \\
        &  TPR@5\%FPR & \bf{0.569} & 0.502 & 0.581 & 0.521& 0.543 \\
        \bottomrule
    \end{tabular}}
    \caption{VeilProbe's performance w.r.t the volume of text-to-suffix mapping features.}
    \label{tab: volume}
\end{table}

\section{Conclusion}
In this paper, we propose VeilProbe, the first automatic pre-training text detection framework for black-box LLMs. We propose a novel sequence-to-sequence mapping model that can effectively capture the Text-to-Suffix mapping pattern and infer latent membership features. Given the scarcity of ground-truth labeled pre-training and non-training samples, we devise a prototype-based classifier to identify pre-training texts and mitigate the overfitting issue. Comprehensive experiments on three widely used datasets demonstrate that our framework significantly outperforms existing baselines. Notably, VeilProbe operates automatically, without any human intervention throughout the entire detection process.

\section*{Acknowledgments}
This work was supported by the Natural Science Foundation of China (No. 62372057).

\bibliographystyle{named}
\bibliography{ijcai24}

\newpage
\appendix

\section{Baseline Details}\label{app: baseline}
We select eight grey-box detection methods and two black-box detection methods as our baselines.

\begin{itemize}
    \item \textbf{Perplexity} \cite{carlini2021extracting}. It is the basic detection method based on perplexity (PPL) by averaging the log-likelihood of all tokens in a sequence.
    \item \textbf{Lowercase} \cite{carlini2021extracting}. It uses the ratio of the perplexity on the sample before and after lowercasing membership scores.
    \item \textbf{Zlib} \cite{carlini2021extracting}. It uses the sequence's zlib compression size to calibrate the sequence's loss as membership scores, which are computed as the number of bits for the sequence when compressed with the zlib library.
    \item \textbf{Neighbor} \cite{mattern-etal-2023-membership}. It computes the membership scores by using the average loss of neighbor sentences with close semantic meaning instead of using a reference model.
    \item \textbf{Min-K\% Prob} \cite{shi2024detecting}. It computes the membership scores using the k\% of tokens with the lowest log-likelihoods, rather than averaging all as in perplexity.
    \item \textbf{Min-K\% Prob ++}  \cite{zhang2024mink}. It adds normalization upon Min-K\% Prob by leveraging the statistics available with the target LLM itself as a reference.
    \item \textbf{DC-PDD}~\cite{zhang2024pretrainingdatadetectionlarge}. It computes a calibrated detection score by using cross-entropy between token probability and reference corpus frequency distributions. This approach addresses the issue that Min-K\% Prob tends to misclassify non-training texts containing high-frequency words.
   \item\textbf{FeatureAgg}~\cite{maini2024llmdatasetinferencedid,puerto2024scalingmembershipinferenceattacks}. It is applied to dataset inference. Here, we extract its main methodology separately to serve as a baseline. It selectively combines the membership scores as the aggregated feature. Then the feature is fed into a linear model trained by some labeled samples to infer membership status.

    \item \textbf{Name-cloze}~\cite{chang2023speakmemoryarchaeologybooks}. It is a black-box method of designing a cloze task for each text. By analyzing the differences in the target LLM's responses to pre-training texts and non-training texts, it can infer membership status. For pre-training texts, the target LLM tends to complete the original words.
    
    \item \textbf{DE-COP} \cite{duarte2024decopdetectingcopyrightedcontent}. It is a black-box detection method that designs multiple-choice questions to let the target LLM answer. The accuracy of the target LLM selecting the correct answer (verbatim original sequence) serves as the detection criterion. 
\end{itemize}

\section{Additional Implementation Details}\label{app: settings}
\noindent\textbf{Baselines}. For Min-K\% and Min-K\% ++ baselines, we followed the previous experimental setup~\cite{shi2024detecting,zhang2024mink}, setting $k=20$. For Neighbor baselines, the number of neighbors is set to $10$, which is consistent with~\cite{zhang2024mink}. For FeatureAgg~\cite{maini2024llmdatasetinferencedid}, we follow the settings provided in the author's demo code and use the following features as input: 
\texttt{ppl}, \texttt{k\_min\_0.05}, \texttt{k\_min\_0.1}, \texttt{k\_min\_0.2}, \texttt{k\_min\_0.3}, \texttt{k\_min\_0.4}, \texttt{k\_min\_0.5}, \texttt{k\_min\_0.6}, \texttt{k\_max\_0.05}, \texttt{k\_max\_0.1}, \texttt{k\_max\_0.2}, \texttt{k\_max\_0.3}, \texttt{k\_max\_0.4}, \texttt{k\_max\_0.5}, \texttt{k\_max\_0.6}, \texttt{zlib\_ratio}.
Similar to our work, since FeatureAgg also relies on ground-truth labeled samples, we set the number of labeled samples to be the same as our VeilProbe, approximately 50 samples for training the linear model.

For existing black-box baselines, since the code for name-cloze is not publicly available, we manually constructed a set of proper nouns as blanks based on the ideas presented in the original paper and allow the LLM to complete them. For DE-COP, some results have been disclosed by the original authors, and we directly present these disclosed results in our main table. For the portions where results are not disclosed, we reproduced them using the provided code.

\noindent\textbf{Our VeilProbe}.  
 In our experimental setup, regardless of whether the LLMs are open-source or closed-source, we assume access is limited to the input and output text, with no use of any hidden information. For each text to be detected, three suffixes are generated by the target LLM. We set the maximum completion suffix length to 512 tokens and perturb Top-10\% of the important tokens in one sequence in the key token perturbation module. The $p$-value significance threshold is set at  \{0.001, 0.01, 0.05, 0.1\} to select the critical perturbation calibration features.  We randomly sample approximately 50 ground-truth samples per dataset to train the prototype-based classifier, with the remaining samples serving as the texts to be detected.
 
Furthermore, for WikiMIA main results in Section~\ref{sec:mainresults}, the results in the main table are obtained by averaging the AUC and TPR@5\%FPR scores across sequence lengths of 32, 64, and 128. The exception is DE-COP, for which only the results at a sequence length of 128 are reported due to cost considerations.
For document-level detection on the 165 books and 50 papers, all methods follow the previous work~\cite{duarte2024decopdetectingcopyrightedcontent}, which use the averaged detection score of the sequences in a document to obtain the AUC and TPR@5\%FPR scores.

\section{Additional Experiments}

\subsection{Common Intuitive Indicators}\label{app:intuitive}
Only the input and output texts are available in the black-box setting. Here, we provide two intuitive indicators that may assist in inferring membership status. (i) \textit{Consistency}: inspired by the previous work~\cite{dong2024generalizationmemorizationdatacontamination,abbasi-yadkori2024to}, we consider that, given a text to be detected, if the text is a member of the target LLM's pre-training data, the multiple output suffixes should be nearly consistent because the LLMs are capable of retaining the knowledge they have seen. (ii) \textit{Semantic Shift}: Intuitively, for training texts, the LLM generates suffixes closely aligned with observed patterns in the training corpus, enabling the LLM to leverage learned associations for semantically consistent completions. In contrast, for non-training texts, the generation of suffixes relies more heavily on the generalization capability of the LLM, exhibiting greater divergence and potentially introducing semantic shifts.

\begin{figure}[ht]
  \centering
  \begin{subfigure}[b]{0.47\linewidth}
    \centering
    \includegraphics[width=\linewidth]{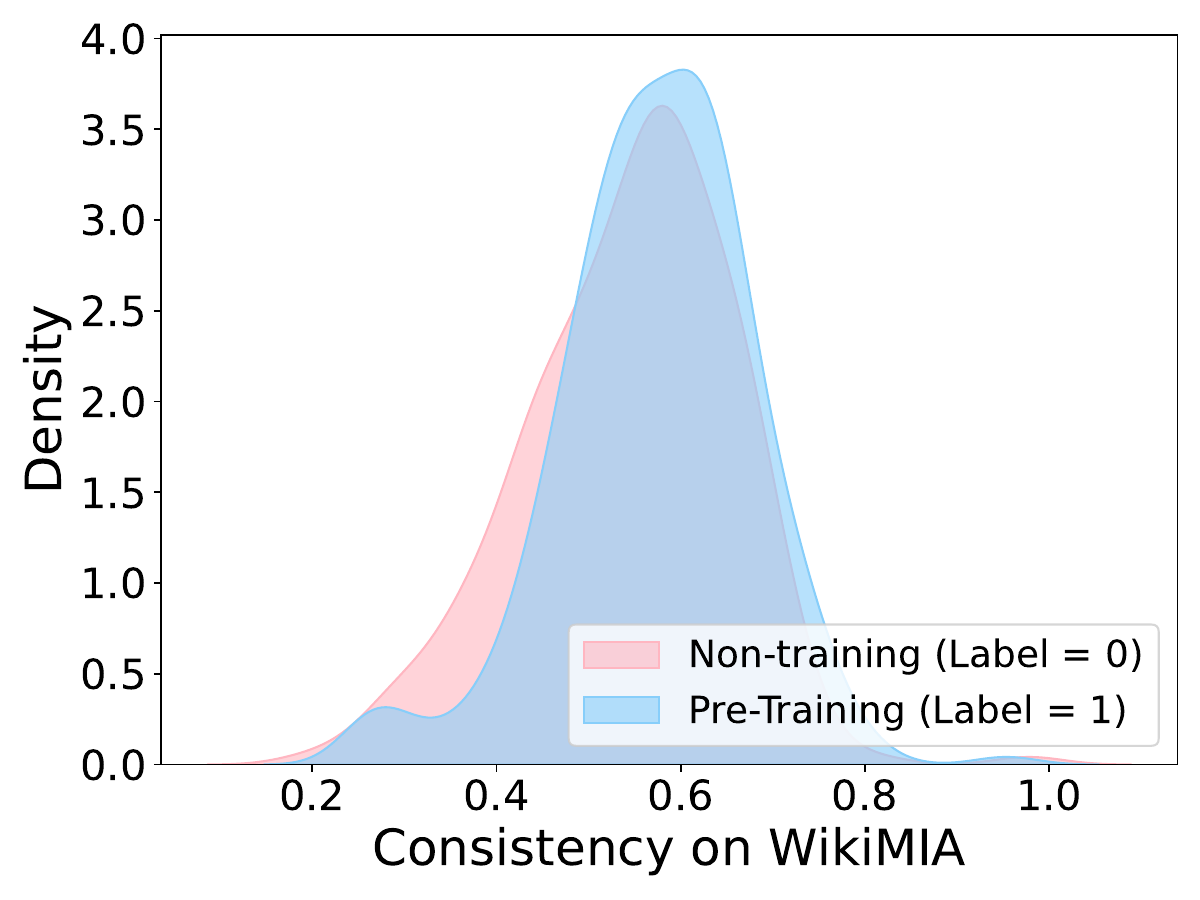}
    \caption{Wiki Consistency}
    \label{fig:wikiconsistency}
  \end{subfigure}
  \hspace*{0.02\linewidth} 
  \begin{subfigure}[b]{0.47\linewidth}
    \centering
    \includegraphics[width=\linewidth]{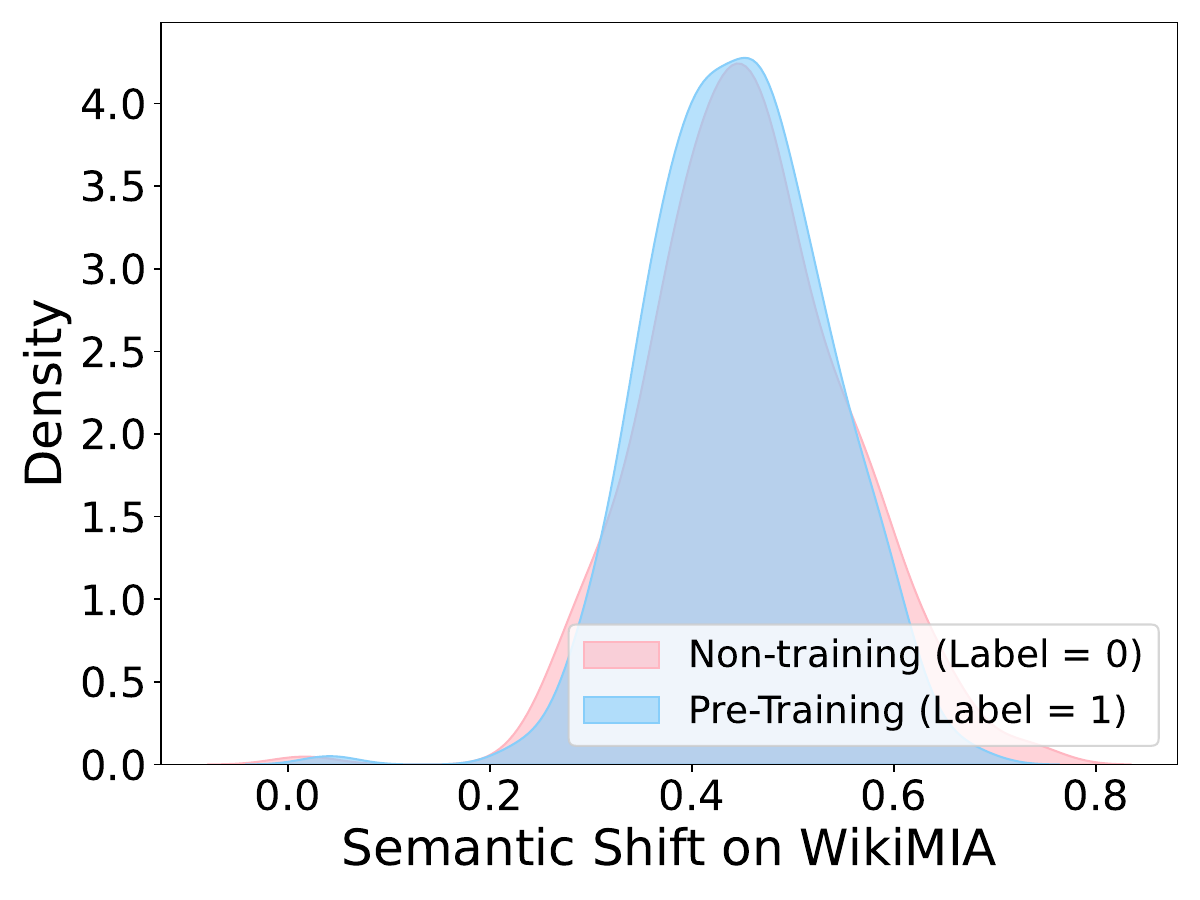}
    \caption{Wiki Shift}
    \label{fig:wikishift}
  \end{subfigure}


  \begin{subfigure}[b]{0.47\linewidth}
    \centering
    \includegraphics[width=\linewidth]{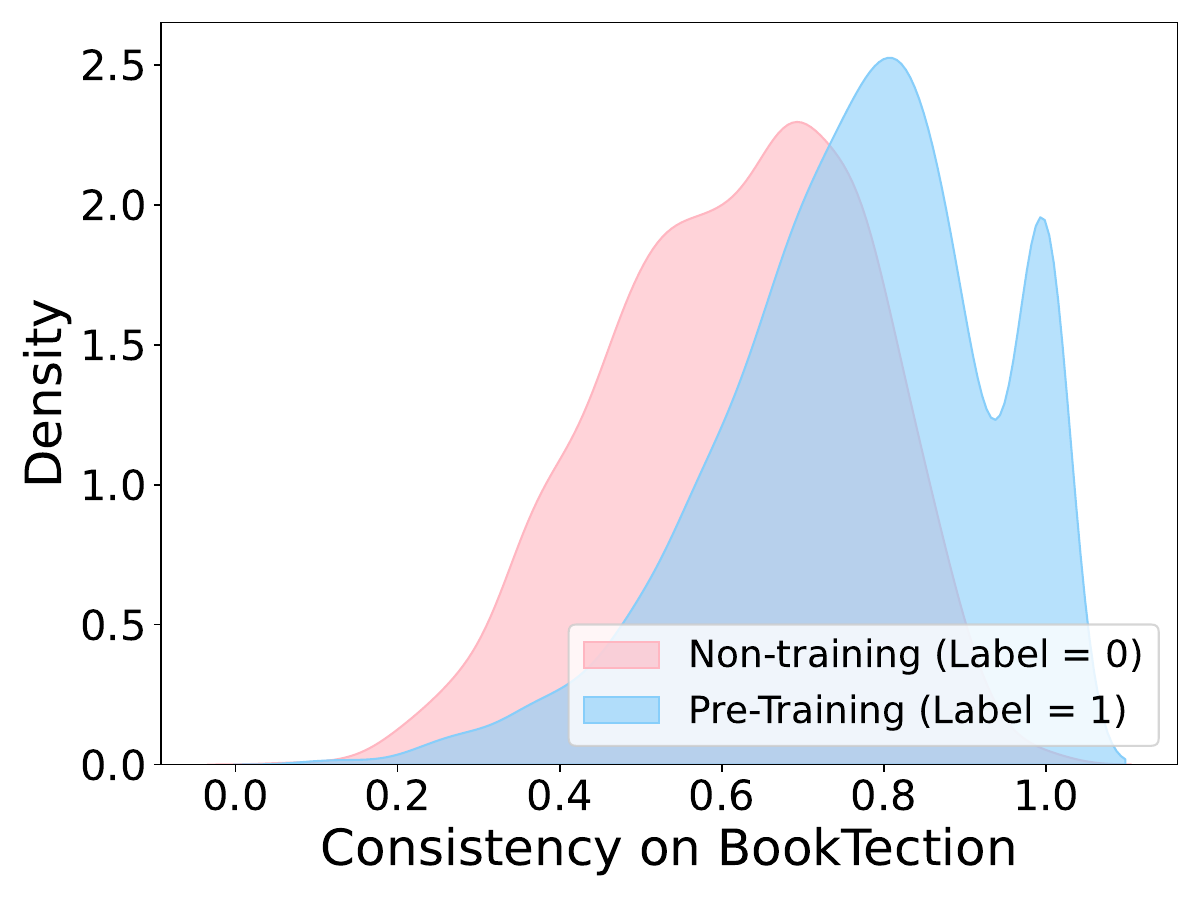}
    \caption{Book Consistency}
    \label{fig:bookconsistency}
  \end{subfigure}
  \hspace*{0.02\linewidth}
  \begin{subfigure}[b]{0.47\linewidth}
    \centering
    \includegraphics[width=\linewidth]{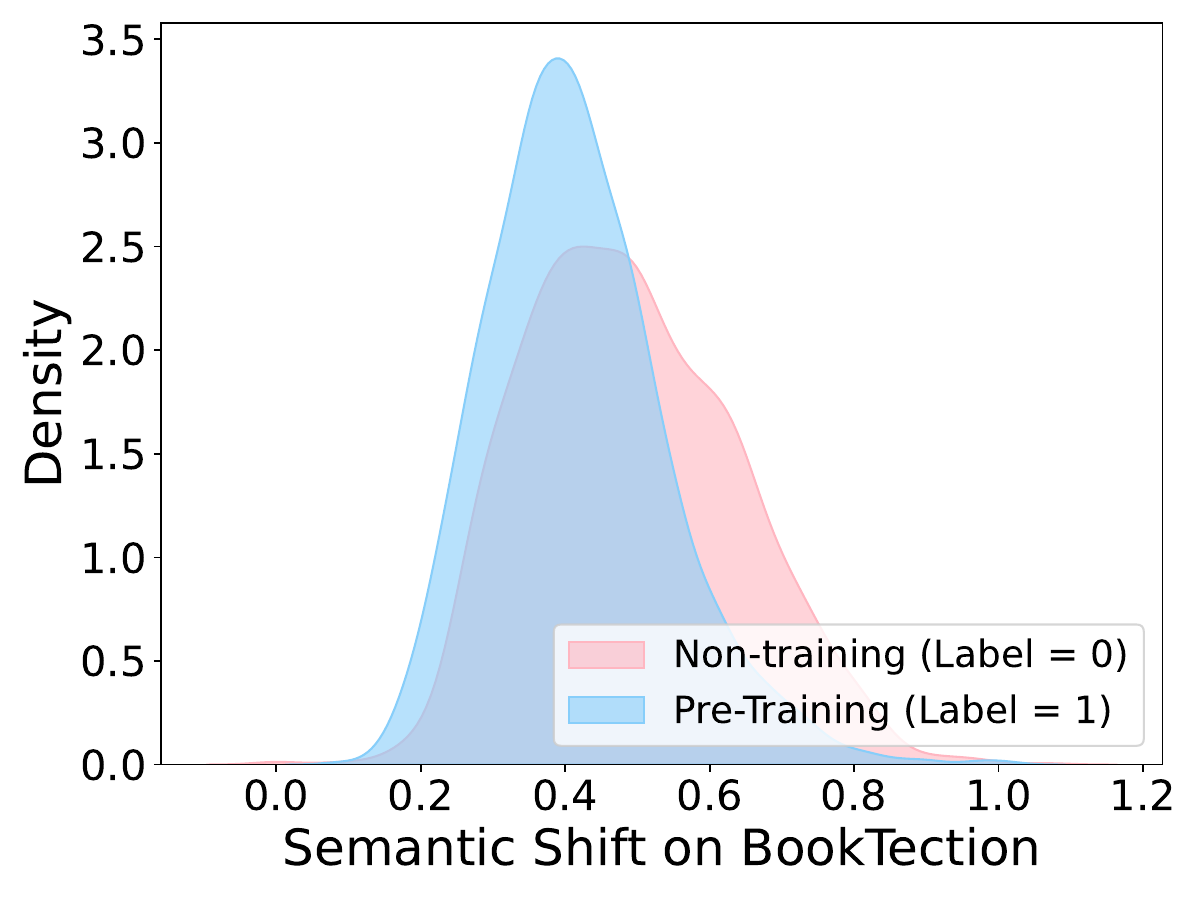}
    \caption{Book Shift}
    \label{fig:bookshift}
  \end{subfigure}


  \begin{subfigure}[b]{0.47\linewidth}
    \centering
    \includegraphics[width=\linewidth]{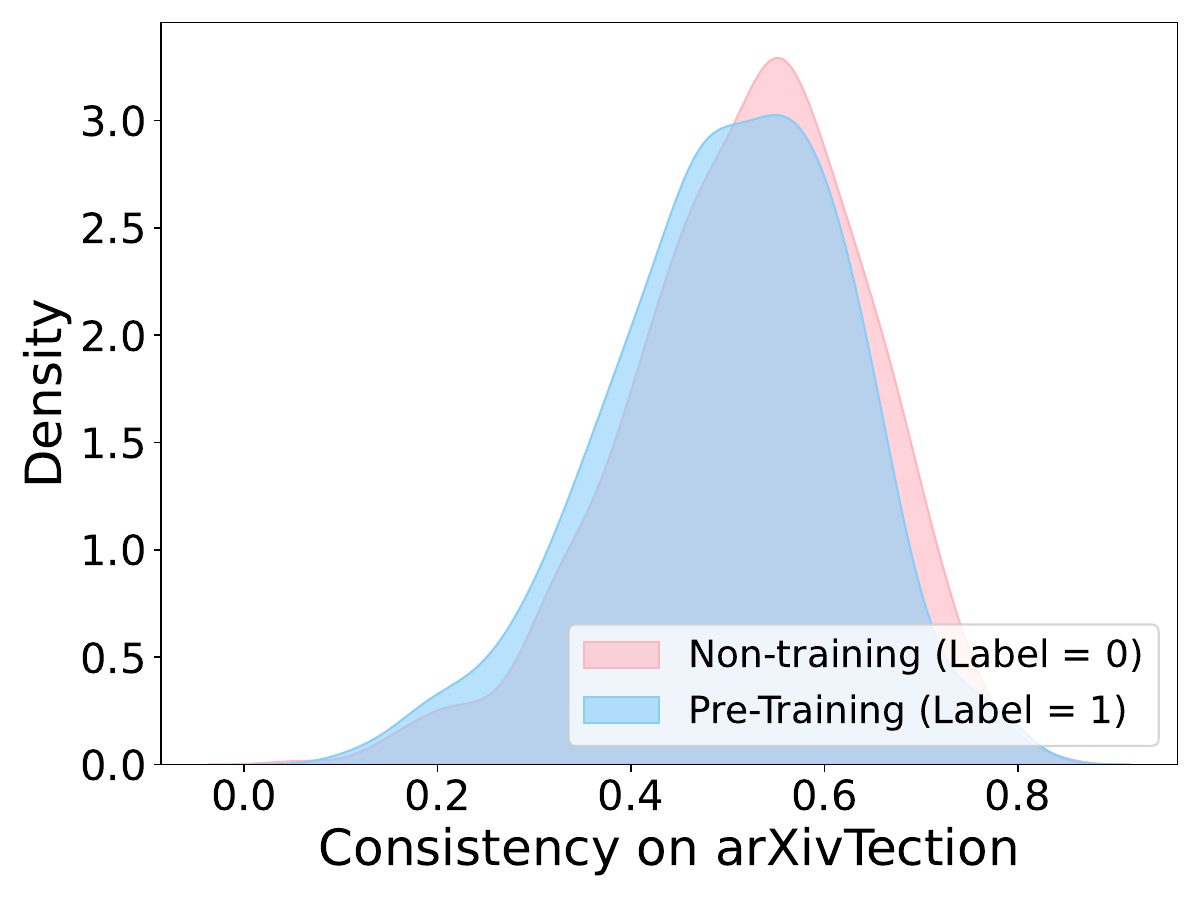}
    \caption{Arxiv Consistency}
    \label{fig:arxivconsistency}
  \end{subfigure}
  \hspace*{0.02\linewidth}
  \begin{subfigure}[b]{0.47\linewidth}
    \centering
    \includegraphics[width=\linewidth]{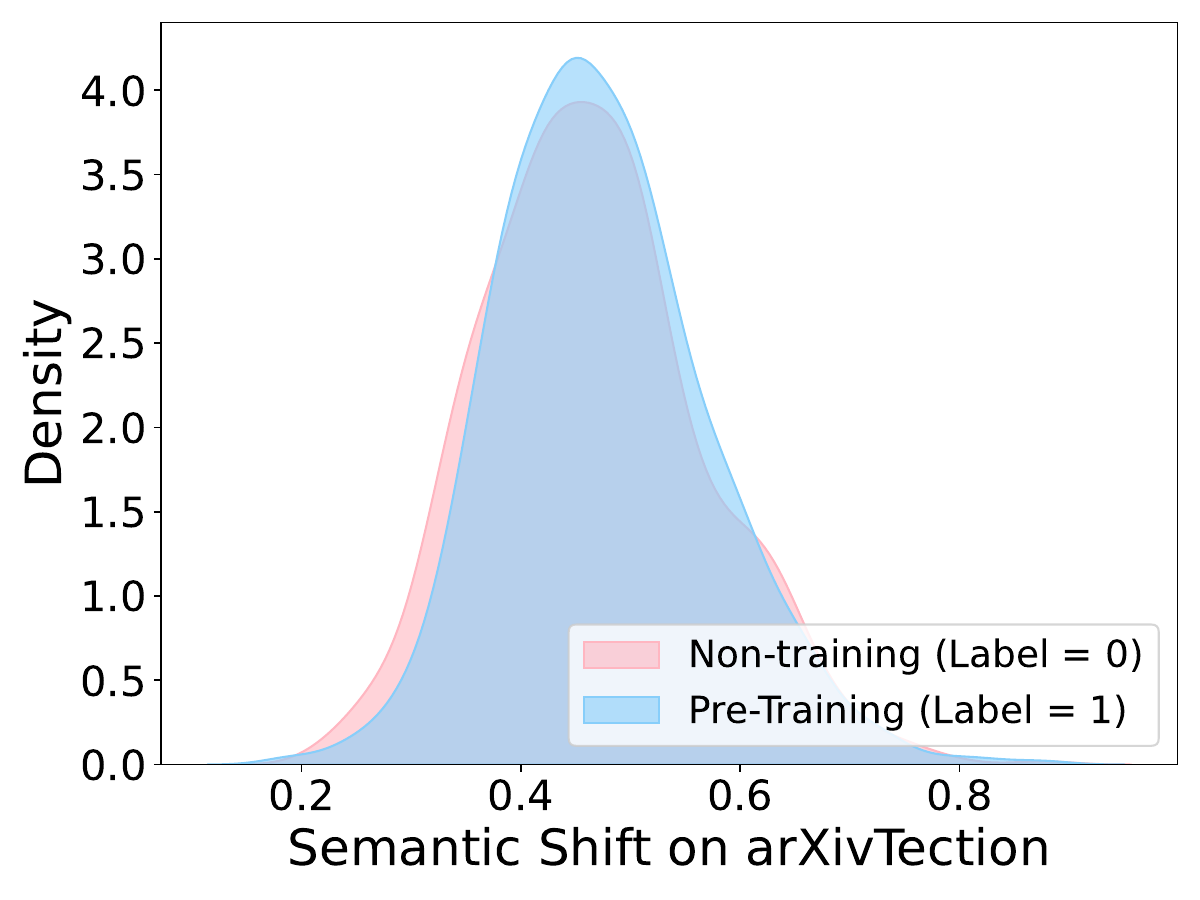}
    \caption{Arxiv Shift}
    \label{fig:arxivshift}
  \end{subfigure}

  \caption{Density plots for consistency and semantic shift across different datasets.}
  \label{fig:density_plots}
\end{figure}

Figure~\ref{fig:density_plots} shows the visualization of the distribution of training and non-training texts on WikiMIA, BookTection, and arXivTection with those two intuitive indicators mentioned earlier. We observe that while BookTection exhibits a subtle distinction, the distributions of pre-training texts and non-training texts on WikiMIA and arXivTection show minimal differentiation. Moreover, in the arXiv domain, some specific samples diverge from the assumptions that serve as the basis for the intuitive indicators.
We also calculate the specific AUC scores, as shown in Table~\ref{tab:intuitive}. The results indicate that common intuitive indicators are often inadequate as distinguishable features. Therefore, it is necessary to infer additional hidden membership features beyond simple textual content.

\begin{table}[ht]
\centering
\resizebox{0.9\columnwidth}{!}{%
\begin{tabular}{cccc}
\toprule
\textbf{Indicator} & \textbf{WikiMIA} & \textbf{BookTection} & \textbf{arXivTection} \\
\hline
Consistency& 0.572 &0.744& 0.451\\
Semantic Shift &0.522 & 0.649& 0.529\\
\bottomrule
\end{tabular}}
\caption{AUC scores on three datasets when using consistency and semantic shift as intuitive indicators.}
\label{tab:intuitive}
\end{table}

\begin{figure}[t]
  \includegraphics[width=1\columnwidth]{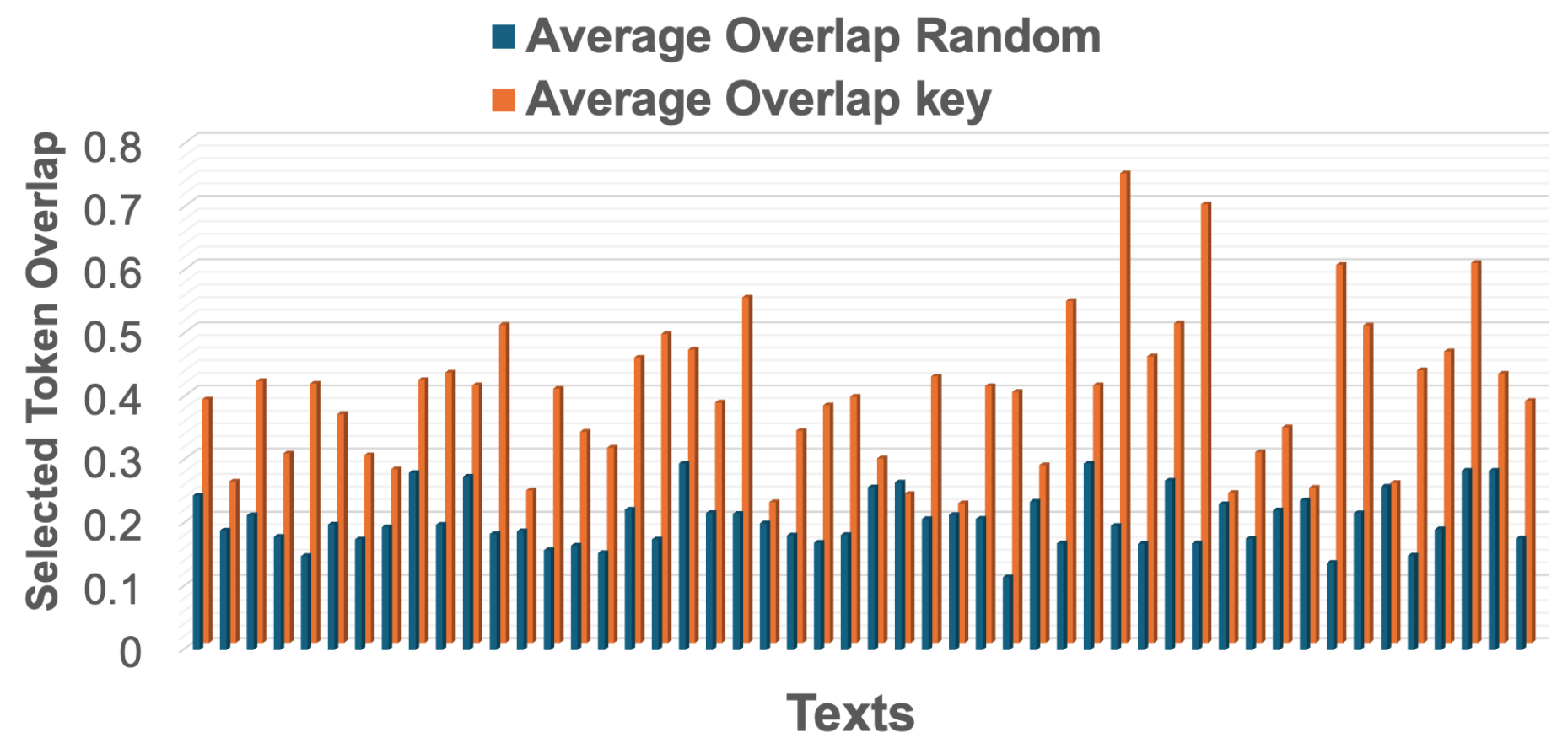}
  \caption{Comparison of selected token overlap rates for 50 texts between random token selection and key token selection. For each text, the tokens selected by using multiple proxy LLMs exhibit a higher overlap compared to random selections, indicating that the feature attribution 
 technique on multiple LLMs shares a specific preference for consistently attributing certain tokens within the texts.}
  \label{fig: overlap}
\end{figure}

\begin{figure}[t]
  \includegraphics[width=1\columnwidth]{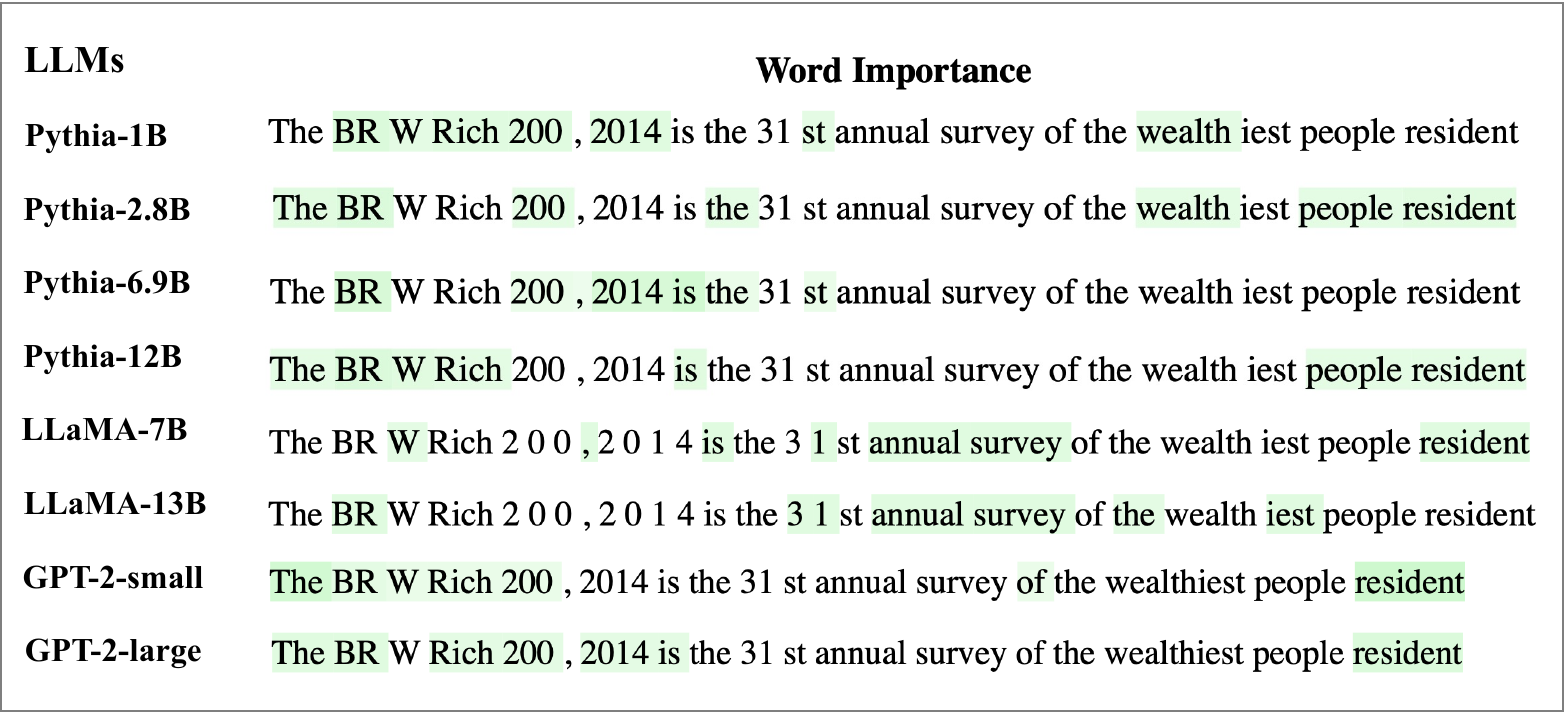}
  \caption{The visualization of token importance across different LLMs.}
  \label{fig: attribute}
\end{figure}

\subsection{Key Token Selection}\label{app: keytoken}
As mentioned in Section~\ref{sec: sampling}, most feature attribution methods rely on hidden information (such as predictions or token probability distributions), which are unsuitable for black-box scenarios where only natural language text outputs are available. In our framework, we employ three small-sized open-source LLMs as proxy LLMs to select informative tokens. Technically, GILOT~\cite{ligilot} is introduced in the key perturbation module in our framework. It utilizes the Optimal Transport to measure distributional changes in generated sequences when each input token is removed, accounting for token similarity to accurately estimate feature attribution in LLMs. 

In this section, we aim to investigate whether the proxy LLMs demonstrate similar preferences for key token selection in feature attribution tasks. The potential reasons for this phenomenon are as follows: First, the semantic and syntactic structures inherent in natural language often exhibit internal consistency, which leads different LLMs to focus on similar words in the prefix when generating the suffix. For example, in a prefix such as \textit{``Please narrate a story about the Olympics,”} the LLM is likely to focus on verbs like \textit{``narrate"} and proper nouns such as \textit{``Olympics"}. Second, the overlap in training corpora across different LLMs likely contributes to this effect. Shared training data can lead models to develop similar tendencies in focusing on and utilizing high-frequency co-occurring words. For example, sentences containing the term \textit{``copyright”} often predictably follow with phrases such as \textit{``legal disputes"} or \textit{``rights protection”}. Third, when LLMs with the same families are selected, the informative tokens they focus on within the prefix are likely to exhibit greater consistency.

We conducted visualization analysis on several LLMs, such as Pythia-1B, Pythia-1.4B, Pythia-6.9B, Pythia-12B, LLaMA-7B, LLaMA-13B, GPT2-small and GPT2-large. As illustrated in Figure~\ref{fig: overlap}, the horizontal axis represents the 50 selected samples, while the vertical axis indicates the overlap rate of elements in the informative token sets chosen by various LLMs.
For a text, the selected token overlap of feature attribution based on the proxy LLMs is higher than just random selection every time. This indicates that using different proxy LLMs to select key tokens would exhibit a bias or preference for certain tokens. Furthermore, Figure~\ref{fig: attribute} presents the visualization results of token importance across different LLMs.
In our experiments, the proxy LLM is configured in three different sizes of the GPT-2 family. By integrating the importance scores obtained from these three small-sized GLMs, we aggregate the scores and select the top-$10$\% key tokens for perturbation operations.

\subsection{Detailed Performance on WikiMIA}
In this section, we present the experimental results on WikiMIA across different lengths in detail. The AUC scores and TPR@5\%FPR scores are shown in Table~\ref{tab:appendixwikiauc} and Table~\ref{tab:appendixwikitpr}. 

\begin{table}[h]
\centering
\scriptsize 
\resizebox{1\columnwidth}{!}{%
\begin{tabular}{cccccc}
\toprule
\textbf{Len.} & \textbf{Method} & \textbf{Pythia-6.9B} & \textbf{LLaMA-13B} & \textbf{NeoX-20B} \\
\midrule
\multirow{6}{*}{32} 
& PPL      & 0.636 & 0.676 & 0.687 \\
& Lowercase & 0.617 & 0.632 & 0.667 \\
& Zlib      & 0.641 & 0.678 & 0.689 \\
& Neighbor  & 0.655 & 0.658 & 0.702 \\
& Min-K\%   & 0.663 & 0.680 & 0.718 \\
& Min-K\%++ & 0.703 & 0.848 & 0.751 \\
& FeatureAgg & 0.647 & 0.651 & 0.718 \\
& Name-cloze& 0.535&0.525& 0.551 \\
& \textbf{VeilProbe} & \textbf{0.930} & \textbf{0.932} & \textbf{0.931} \\
\midrule
\multirow{6}{*}{64} 
& PPL      & 0.617 & 0.644 & 0.674 \\
& Lowercase & 0.577 &  0.610 & 0.651 \\
& Zlib      & 0.618 & 0.659 & 0.687\\
& Neighbor  & 0.633 & 0.641 & 0.671 \\
& Min-K\%   & 0.632 & 0.669 & 0.735 \\
& Min-K\%++ & 0.672 & 0.856 & 0.760 \\
& FeatureAgg & 0.632 & 0.615 & 0.717 \\
& Name-cloze& 0.536&0.558& 0.524 \\
& \textbf{VeilProbe} & \textbf{0.937} & \textbf{0.911} & \textbf{0.947} \\
\midrule
\multirow{6}{*}{128} 
& PPL      & 0.651 & 0.678 & 0.707 \\
& Lowercase & 0.605 & 0.564& 0.680 \\
& Zlib      & 0.676 & 0.697 & 0.723 \\
& Neighbor  & 0.660 & 0.662 & 0.696 \\
& Min-K\%   & 0.695 & 0.715 & 0.756\\
& Min-K\%++ & 0.697 & 0.838 & 0.754 \\
& DC-PDD & 0.698 & 0.697 & 0.766 \\
& FeatureAgg & 0.673& 0.669 & 0.748\\
& Name-cloze& 0.600& 0.519 & 0.500 \\
& DE-COP & 0.512& 0.512 & 0.531\\
& \textbf{VeilProbe} & \textbf{0.871} &\textbf{0.864}& \textbf{0.870} \\
\bottomrule
\end{tabular}%
}
\caption{Detailed AUC scores across different methods on WikiMIA across varying sequence lengths.}
\label{tab:appendixwikiauc}
\end{table}

\begin{table}[H]
\centering
\scriptsize 
\resizebox{1\columnwidth}{!}{%
\begin{tabular}{cccccc}
\toprule
\textbf{Len.} & \textbf{Method} & \textbf{Pythia-6.9B} & \textbf{LLaMA-13B} & \textbf{NeoX-20B} \\
\midrule
\multirow{6}{*}{32} 
& PPL      &  0.142  &  0.139& 0.199 \\
& Lowercase & 0.106 & 0.096 & 0.181 \\
& Zlib      & 0.163& 0.116 & 0.199  \\
& Neighbor  &0.165  & 0.116 &0.222  \\
& Min-K\%   &  0.178 & 0.189  &0.279  \\
& Min-K\%++ &  0.171 &  0.385  & 0.194  \\
& FeatureAgg & 0.166 & 0.113& 0.268 \\
& Name-cloze& 0.091&0.066& 0.090 \\
& \textbf{VeilProbe} & \textbf{0.694} & \textbf{0.742}&\textbf{0.722} \\
\midrule
\multirow{6}{*}{64} 
& PPL      &  0.134  & 0.113& 0.130  \\
& Lowercase & 0.116 & 0.116 & 0.155 \\
& Zlib      &  0.162  & 0.127& 0.166 \\
& Neighbor  & 0.109  &0.102& 0.130 \\
& Min-K\%   & 0.190 & 0.172 &0.204  \\
& Min-K\%++ &0.261 &0.341 &0.204\\
& FeatureAgg & 0.178 & 0.120 & 0.240 \\
& Name-cloze& 0.083&0.116& 0.090 \\
& \textbf{VeilProbe} & \textbf{0.731} & \textbf{0.543} & \textbf{0.711}\\
\midrule
\multirow{6}{*}{128} 
& PPL      & 0.144 & 0.216 & 0.180 \\
& Lowercase & 0.130 & 0.158 & 0.120\\
& Zlib      & 0.209& 0.187 & 0.223 \\
& Neighbor  & 0.108 &0.129& 0.158 \\
& Min-K\%   & 0.180 & 0.201& 0.216\\
& Min-K\%++ & 0.201 &\textbf{0.381} & 0.245 \\
& DC-PDD & 0.245& 0.230 & 0.317 \\
& FeatureAgg & 0.153 & 0.126 & 0.180 \\
& Name-cloze & 0.106& 0.027 & 0.058 \\
& DE-COP & 0.061 & 0.064 & 0.092\\
& \textbf{VeilProbe} & \textbf{0.451} & 0.378 &\textbf{0.441}  \\
\bottomrule
\end{tabular}%
}
\caption{Detailed TPR@5\%FPR scores across different methods on WikiMIA across varying sequence lengths.}
\label{tab:appendixwikitpr}
\end{table}

\subsection{Detailed Comparative Experiments on BookTection}\label{app: book}
Tables~\ref{tab: appbookauc} and \ref{tab: appbooktpr} present the results of comparative experiments conducted on document-level detection across 165 books in the BookTection.

\begin{table}[ht]
\centering
\resizebox{0.8\columnwidth}{!}{%
\begin{tabular}{cccc}
\toprule
\textbf{Method} & \textbf{Mistral-7B} & \textbf{LLaMA2-7B} & \textbf{ChatGPT} \\
\hline
PPL & 0.785 &0.838&  \ding{55}\\
Lowercase &0.894 & 0.913& \ding{55}\\
Zlib & 0.537& 0.594 &\ding{55} \\
Neighbor & 0.855 & 0.912 &\ding{55} \\
Min-K\% & 0.799 & 0.856&\ding{55} \\
Min-K\%++ & 0.694& 0.807& \ding{55} \\
FeatureAgg &0.808 & 0.798 &\ding{55} \\
Name-cloze &0.742& 0.733& 0.726\\
DE-COP &0.901& 0.780& \bf{0.990}\\
VeilProbe & \bf{0.996}& \bf{0.998} &0.988 \\
\bottomrule
\end{tabular}}
\caption{AUC scores for detecting on 165 books. Performance of various methods across different target LLMs}
\label{tab: appbookauc}
\end{table}

\begin{table}[ht]
\centering
\resizebox{0.8\columnwidth}{!}{%
\begin{tabular}{cccc}
\toprule
\textbf{Method} & \textbf{Mistral-7B} & \textbf{LLaMA2-7B} & \textbf{ChatGPT} \\
\hline
PPL & 0.362 &0.419&  \ding{55}\\
Lowercase &0.371 & 0.686& \ding{55}\\
Zlib & 0.100& 0.167&\ding{55} \\
Neighbor & 0.581& 0.648 &\ding{55} \\
Min-K\% & 0.371 & 0.429&\ding{55} \\
Min-K\%++ & 0.200& 0.343& \ding{55} \\
FeatureAgg &0.410 & 0.429&\ding{55} \\
Name-cloze &0.400& 0.362& 0.410\\
DE-COP &0.667& 0.524& 0.924\\
VeilProbe &\bf{0.991}& \bf{1.000}&\bf{0.943} \\
\bottomrule
\end{tabular}}
\caption{TPR@5\%FPR scores for detecting on 165 books. Performance of various methods across different target LLMs}
\label{tab: appbooktpr}
\end{table}

\subsection{Detailed Comparation Experiments on ArxivTection}\label{app: arxiv}

Tables~\ref{app: arxivauc} and~\ref{app: arxivtpr} show the results of comparative experiments conducted on document-level detection across 50 papers in the arXivTection.

\begin{table}[H]
\centering
\resizebox{0.8\columnwidth}{!}{%
\begin{tabular}{cccc}
\toprule
\textbf{Method} & \textbf{Mistral-7B} & \textbf{LLaMA2-13B} & \textbf{Claude 2.1} \\
\hline
PPL & 0.894 &0.883&  \ding{55}\\
Lowercase &0.619 &0.494& \ding{55}\\
Zlib & 0.741& 0.707 &\ding{55} \\
Neighbor & 0.582 & 0.613 &\ding{55} \\
Min-K\% & 0.965&0.966&\ding{55} \\
Min-K\%++ & 0.939&0.965& \ding{55} \\
FeatureAgg &\bf{0.989}& 0.978 &\ding{55} \\
Name-cloze &0.503 &0.704& 0.894\\
DE-COP &0.503& 0.503& 0.908\\
VeilProbe &0.978& \bf{0.994} &\bf{0.994}\\
\bottomrule
\end{tabular}}
\caption{AUC scores for detecting on 50 arXiv papers. Performance of various methods across different target LLMs}
\label{app: arxivauc}
\end{table}


\begin{table}[H]
\centering
\resizebox{0.8\columnwidth}{!}{%
\begin{tabular}{cccc}
\toprule
\textbf{Method} & \textbf{Mistral-7B} & \textbf{LLaMA2-13B} & \textbf{Claude 2.1} \\
\hline
PPL & 0.275 &0.250&  \ding{55}\\
Lowercase &0.018 &0.005& \ding{55}\\
Zlib & 0.090&0.047 &\ding{55} \\
Neighbor & 0.005 &0.010 &\ding{55} \\
Min-K\% & 0.770&0.810&\ding{55} \\
Min-K\%++ & 0.450&0.740& \ding{55} \\
FeatureAgg &0.880& 0.920 &\ding{55} \\
Name-cloze &0.000 &0.240& 0.480\\
DE-COP &0.000&0.000&0.480\\
VeilProbe & \bf{0.920}& \bf{0.960} &\bf{0.960} \\
\bottomrule
\end{tabular}}
\caption{TPR5\%FPR scores for detecting on 50 arXiv papers. Performance of various methods across different target LLMs}
\label{app: arxivtpr}
\end{table}

\end{document}